%% file: main.tex
\documentclass[nohyperref]{article}

\usepackage{microtype}
\usepackage{graphicx}
\usepackage{caption}
\usepackage{subcaption}
\usepackage{booktabs} 
\usepackage{algorithm} 
\usepackage{algorithmic}
\usepackage{hyperref}

\usepackage[accepted]{icml2022}

\usepackage{amsmath}
\usepackage{amssymb}
\usepackage{mathtools}
\usepackage{amsthm}

\usepackage[capitalize,noabbrev]{cleveref}

\usepackage{multirow}
\usepackage{multicol}
\usepackage{arydshln}
\usepackage{slashbox}

\usepackage{enumitem}

\theoremstyle{plain}

\theoremstyle{definition}

\theoremstyle{remark}

\usepackage[textsize=tiny]{todonotes}

\newcommand\ghkim[1]{\textcolor[rgb]{0.5,0.5,0.9}{#1}}

\newcommand{\revision}[1]{\textcolor{cyan}{#1}}

\icmltitlerunning{Multi-Level Branched Regularization for Federated Learning}
\begin{document}

\twocolumn[
\icmltitle{Multi-Level Branched Regularization for Federated Learning}



\icmlsetsymbol{equal}{*}

\begin{icmlauthorlist}
\icmlauthor{Jinkyu Kim}{equal,cvlab}
\icmlauthor{Geeho Kim}{equal,cvlab}
\icmlauthor{Bohyung Han}{cvlab,ipai}
\end{icmlauthorlist}

\icmlaffiliation{cvlab}{Computer Vision Laboratory, Department of Electrical and Computer Engineering \& ASRI, Seoul National University, Korea}
\icmlaffiliation{ipai}{Interdisciplinary Program of Artificial Intelligence, Seoul National University, Korea}

\icmlcorrespondingauthor{Bohyung Han}{bhhan@snu.ac.kr}

\icmlkeywords{Machine Learning, ICML}

\vskip 0.3in
]



\printAffiliationsAndNotice{\icmlEqualContribution} 
%

\input{sections/abstract.tex}
\input{sections/introduction.tex}
\input{sections/related_work.tex}

\input{sections/preliminary.tex}
\input{sections/method.tex}

\input{sections/experiments.tex}
\input{sections/conclusion.tex}

\input{sections/acknowledgments}


\bibliographystyle{icml2022}
\bibliography{egbib}

\input{sections/appendix.tex}


\end{document}

%% file: sections/abstract.tex

\begin{abstract}

A critical challenge of federated learning is data heterogeneity and imbalance across clients, which leads to inconsistency between local networks and unstable convergence of global models.
To alleviate the limitations, we propose a novel architectural regularization technique that constructs multiple auxiliary branches in each local model by grafting local and global subnetworks at several different levels and that learns the representations of the main pathway in the local model congruent to the auxiliary hybrid pathways via online knowledge distillation.
The proposed technique is effective to robustify the global model even in the non-iid setting and is applicable to various federated learning frameworks conveniently without incurring extra communication costs. 
We perform comprehensive empirical studies and demonstrate remarkable performance gains in terms of accuracy and efficiency compared to existing methods.
The source code is available in our project page\footnote{\url{http://cvlab.snu.ac.kr/research/FedMLB}}.

\end{abstract}

%% file: sections/introduction.tex

\section{Introduction}
\label{sec:introduction}

Training deep neural networks typically relies on centralized algorithms, where computing resources and training data are located within a single server.
Recently, to deal with large-scale models and/or distributed data, the learning frameworks based on multiple remote machines become widespread in machine learning research and development.
Federated learning~\cite{mcmahan2017communication} is a unique distributed learning framework that takes advantage of computing resources and training data in clients, typically edge devices, which is helpful to secure data privacy but often suffers from insufficient computing power, low communication bandwidth, and extra battery consumption.
As practical solutions to handle the challenges, each edge device often runs a small number of iterations and minimizes the communication rounds with the central server.

\begin{figure}[t]
\centering
\includegraphics[width=1\linewidth]{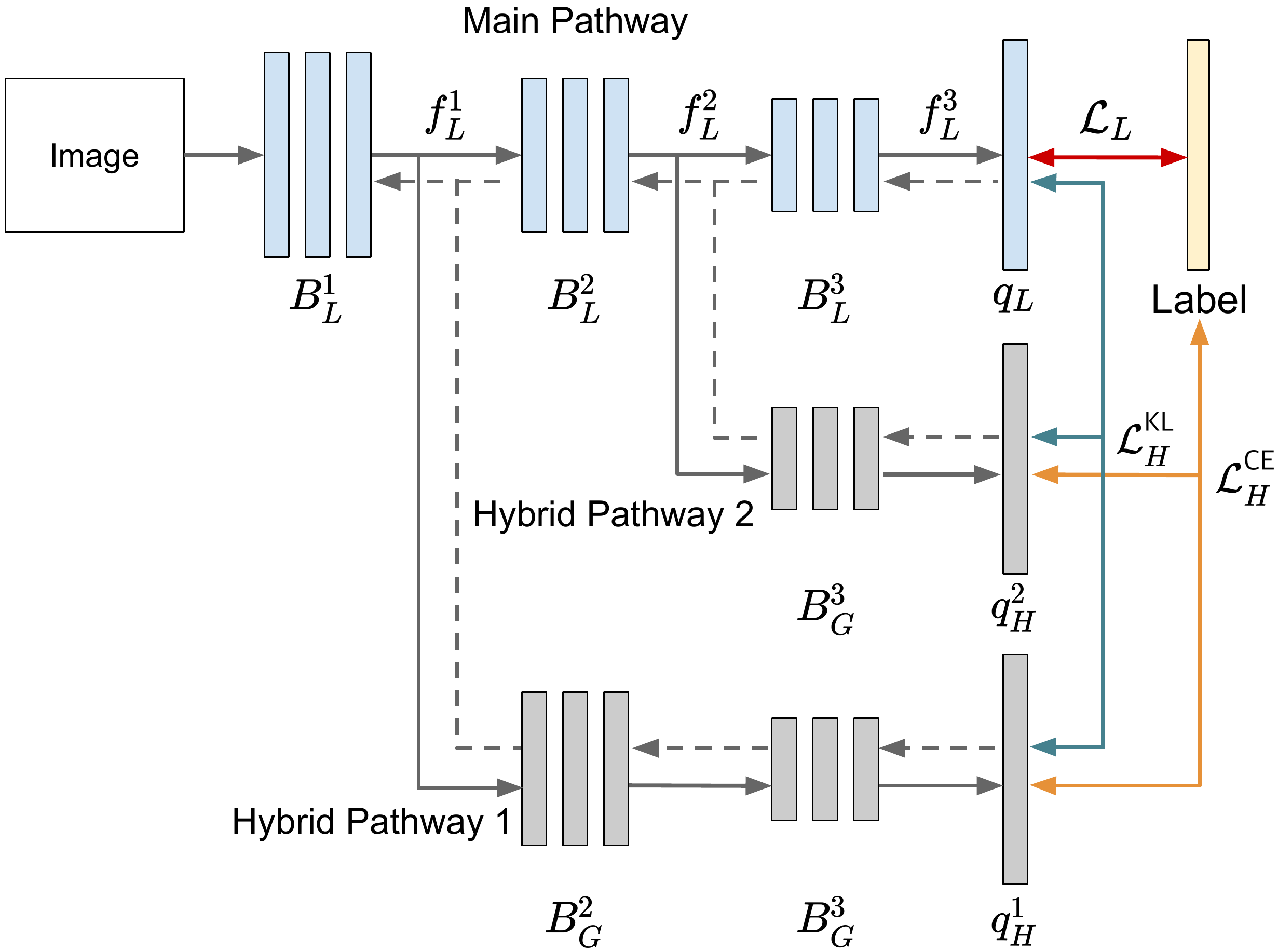}
\caption{Illustration of the local network for the proposed multi-level branched regularization framework. $B_L$ and $B_G$ denote network blocks extracted from local and global networks, respectively, and their superscripts indicate block indices. $q_L$ and $q_H$ indicate softmax outputs of the main pathway and hybrid pathways, respectively.
Solid and dashed gray lines represent forward and backpropagation flows, respectively.
For training, the standard cross-entropy losses are applied to all branches and the KL-divergence losses between the main pathway and the rest of the pathways are employed for regularization.
Note that we update the parameters in the main pathway only, which are illustrated in sky blue color in this figure.}
\label{fig:framework}
\end{figure}

FedAvg~\cite{mcmahan2017communication}, the standard optimization method of federated learning, maintains the global model in the server by aggregating the local models trained independently in multiple clients.
Each client utilizes its own dataset to update the local model instead of sharing the data with the server or other clients, sends the locally optimized model to the server for aggregation, and downloads the updated global model for the next stage.
%
FedAvg works well when the individual datasets of the clients are iid and the participation rate of the clients is high, but it struggles with low convergence speed otherwise. 
This is because, when the data distributions of individual clients are different from the global distribution, local updates are prone to drift and increase divergence with respect to the global model~\cite{zhao2018federated, karimireddy2019scaffold}.

Our goal is to make each client preserve the latest global representations and prevent model drift caused by independent local updates.
Meanwhile, we make the server learn client-specific knowledge by a simple aggregation of local networks and achieve a high-performance model well-suited for all clients, eventually. 
To mitigate the discrepancy between the local models updated by heterogeneous datasets, we propose a novel architectural regularization technique via knowledge distillation that grafts the local and global subnetworks and constructs multi-level branches.
The auxiliary branches reduce the deviation of the representations in the local models from the feature space of the global model.
To this end, a participating client first modularizes the downloaded global network into
multiple blocks.
The client trains the main pathway of the local model with the parameters corresponding to the global subnetworks fixed, while the output representations of the hybrid pathways are made similar to those in the main pathway using additional regularization terms based on knowledge distillation.
Our regularization approach is unique compared to the methods based on the standard knowledge distillation because it constrains the representations of the main pathway in the local model using the on-the-fly outputs of the hybrid pathways.
This idea is motivated by a recent knowledge distillation approach that aims to learn a student-friendly teacher network~\cite{park2021learning}.
Figure~\ref{fig:framework} illustrates our main idea.

The main contributions of this work are as follows:
 \begin{itemize} 
\item[$\bullet$]We propose a simple but effective regularization technique to reduce the drift problem of a local model, via online knowledge distillation between the main pathway and the multiple hybrid pathways reflecting the global representations partially.
\item[$\bullet$]Our approach is robust to typical challenges of federated learning including data heterogeneity and low client participation rate, and is applicable to various federated learning frameworks.
\item[$\bullet$]The proposed method requires no additional communication cost and spends no extra memory to store the history of local or global states and the auxiliary information.
\item[$\bullet$]We demonstrate through extensive experiments that our architectural regularization technique improves accuracy and convergence speed consistently compared to the state-of-the-art federated learning algorithms.
 \end{itemize}

The organization of this paper is as follows. 
We first discuss existing work related to the optimization in federated learning and review the basic concept with a simple solution in Section~\ref{sec:related} and \ref{sec:preliminaries}, respectively.
Section~\ref{sec:proposed} describes the proposed federated learning framework, and Section~\ref{sec:experiments} demonstrates the effectiveness of the proposed approach via extensive experiments.
Finally, we conclude this paper in Section~\ref{sec:conclusion}.

%% file: sections/related_work.tex

\section{Related Work}
\label{sec:related}
Federated learning~\cite{mcmahan2017communication} is a distributed learning framework with the characteristics such as non-iid client data, data privacy requirement, massive distribution, and partial participation.
Although FedAvg provides a practical solution for the issues, it still suffers from the heterogeneity of data across clients~\cite{zhao2018federated}.
On the theoretical side, there exist several works that derive convergence rates with respect to the data heterogeneity~\cite{li2020federated, wang2019adaptive, khaled2019first, li2019convergence, hsieh2020non, wang2020tackling}.

To alleviate the limitations of FedAvg, the local model updates are often regularized to prevent a large deviation from the global model. 
FedProx~\cite{li2020federated} imposes a quadratic penalty over the distance between the server and client parameters
while SCAFFOLD~\cite{karimireddy2019scaffold} and FedDANE~\cite{li2019feddane} employ a form of variance reduction techniques such as control variates.
On the other hand, FedPD~\cite{zhang2020fedpd} and FedDyn~\cite{acar2020federated} penalize each client's risk objective dynamically based on its local gradient.
Some approaches adopt a trick motivated by data augmentation~\cite{yoon2021fedmix}\ghkim{,} or contrastive learning~\cite{li2021model} to ensure the similarity of the representations between the downloaded global model and local networks.
However, these methods typically rely on unrealistic or impractical assumptions such as high participation rates, additional communication costs, or extra memory requirements in clients.

Meanwhile, the server-side optimization techniques have been discussed actively for the acceleration of the convergence.
For example, FedAvgM~\cite{hsu2019measuring} adds a momentum term to speed up training, and FedADAM~\cite{reddi2021adaptive} adopts an adaptive gradient-descent method.
Our approach and these aggregation-based methods are orthogonal, so combining them together can yield additional performance gains.

There is another line of research that utilizes knowledge distillation to tackle data heterogeneity issue  in federated learning.
The algorithms in this category perform knowledge distillation either in the server or clients.
As client distillation methods, FD~\cite{seo2020federated} shares the representations between clients for knowledge distillation with their ensemble features while FedLS-NTD~\cite{lee2021preservation} transfers the knowledge of the global model to local networks except the activations for ground-truth labels.
FedGKD~\cite{yao2021local} employs representations from the ensemble of the historical global models to refine local models, and FedGen~\cite{zhu2021data} learns a global generator to aggregate the local information and distill global knowledge to clients.
For the distillation in the server, FedDF~\cite{lin2020ensemble} utilizes the averaged representations of local models on proxy data for aggregation.
However, these methods require additional communication overhead~\cite{yao2021local, zhu2021data} or auxiliary data~\cite{seo2020federated, lin2020ensemble}.
In particular, as discussed in~\cite{wang2021field}, federated learning algorithms are sensitive to the communication cost and the use of globally-shared auxiliary data should be cautiously performed. 
To the contrary, our approach incurs no additional communication cost and has no requirement of auxiliary data.
The proposed algorithm belongs to the method for local optimization based on knowledge distillation.

%% file: sections/preliminary.tex


\section{Preliminaries}
\label{sec:preliminaries}
This section briefly discusses the concept and procedure of the basic federated learning algorithm.

\subsection{Problem setting and notations}
Given $N$ clients, the goal of the federated learning is to learn a global model $\theta$ that minimizes the average losses of all clients as follows:
\begin{equation}
\label{eq:global_objective}
    \underset{\theta }{\text{argmin}} \left[ \mathcal{L}(\theta) = \frac{1}{N} \sum_{i=1}^N \mathcal{L}_i(\theta) \right],
\end{equation}
where $\mathcal{L}_i(\theta)=\mathbb{E}_{{(x,y)} \sim D_i}[ \mathcal{L}_i(x,y;  \theta)]$ is the loss in the $i^\text{th}$ client given by the expected loss over all instances in the client, denoted by $D_i$.
Note that clients may have heterogeneous data distributions and exchanges of training data are strictly prohibited due to privacy issues. 
%

\subsection{FedAvg algorithm}
FedAvg~\cite{mcmahan2017communication} is a standard solution of federated learning, where the server simply aggregates all the participating client models to obtain the global model.
Specifically, in the $t^\text{th}$ communication round, a central server first sends a global model $\theta^{t-1}$ to each of the clients.
Each client sets its initial parameter $\theta^{t}_{i, 0}$ to $\theta^{t-1}$, \textit{i.e.}, $\theta_{i, 0} = \theta^{t-1}$,  performs $K$ steps of the gradient descent optimization to minimize its local loss, and then returns the resultant model parametrized by $\theta_{i,K}^t$ to the server. 
An updated global model for the next round training is obtained by averaging all the participating local models in the current communication round. 
The local loss of FedAvg at the $k^\text{th}$ local iteration ($k = 1, \dots, K$) is defined by
\begin{equation}
\label{fedavg_loss}
    \mathcal{L}_i(\theta_{i,k}^t) = \mathbb{E}_{{(x,y)} \sim D_i}[\mathcal{L}_i(x,y ; \theta_{i,k}^t)],
\end{equation}
where the cross-entropy loss is typically used for $\mathcal{L}_i$.

Multiple local updates in FedAvg before the aggregation step in the server decreases the communication cost for training apparently.
However, in practice, it typically leads to the so-called client drift issue~\cite{karimireddy2019scaffold}, where the individual client updates are prone to be inconsistent due to overfitting on local client data. 
This phenomenon inhibits FedAvg from converging to the optimum of the average loss over all clients.

%% file: sections/method.tex


\section{Proposed Algorithm: FedMLB}
\label{sec:proposed}
This section describes the details of our approach for federated learning with multi-level branched regularization, referred to as FedMLB, which exploits the representations of multiple hybrid pathways.

\subsection{Overview}
\label{sub:overview}
The main objective of FedMLB is to prevent the representations of the local model from being deviated too much by local updates while accommodating new knowledge from each client with heterogeneous datasets through independent local updates.
We achieve this goal via indirect layer-wise online knowledge distillation using the architecture illustrated in Figure~\ref{fig:framework}.

Although there exist several regularization approaches based on knowledge distillation for federated learning~\cite{lee2021preservation, yao2021local, zhu2021data}, FedMLB is unique in the sense that it constructs multiple hybrid pathways combining the subnetworks of local and global models at various levels and learns the representations of the local model similar to those of the hybrid pathways.
Note that the proposed approach performs knowledge distillation using on-the-fly targets given by the hybrid pathways although the subnetworks of the global model remain fixed.
The regularization using the multiple auxiliary branches plays a critical role to make the individual blocks of the local network aligned well to the matching subnetworks of the global model.
Consequently, local updates in each client allow the local model to be less deviated from the global counterpart, and such a regularization also reduces the variations of the models collected from multiple clients.
We describe the details of the proposed algorithm next.

\subsection{Multi-level hybrid branching}
\label{sub:multi}

To perform the proposed regularization, we first divide the network into $M$ exclusive blocks, which are based on the depths and the feature map sizes of the architecture.
Let $\{B_L^m\}_{m=1}^M$ and $\{B_G^m\}_{m=1}^M$ be the sets of blocks in a local model and the global network, respectively.
The main pathway consists of local blocks $B_L^{1:M}$ and we create multiple hybrid pathways by augmenting a subnetwork in the global model $B_G^{m+1:M}$ to a local subnetwork $B_L^{1:m}$. 
Depending on branching locations, from $1$ to $M-1$, several different hybrid pathways, denoted by $\{ B_L^{1:m}, B_G^{m+1:M} \}_{{m \in  \{1,\dots,M-1\}}}$, are constructed in parallel as illustrated in Figure~\ref{fig:framework}.

The constructed network by multi-level hybrid branching has $M$ pathways altogether for predictions, which includes one main pathway and $M-1$ hybrid pathways.
The softmax output of the main pathway, $q_L(x; \tau )$, is given by
\begin{equation}
 q_L(x; \theta, \tau) = \text{softmax} \left (\frac{f_L(x; \theta)}{\tau} \right), 
\end{equation}
where $x$ is the input of the network, $\theta$ is the model parameters, $\tau$ is the temperature of the softmax function, and $f_L(\cdot ; \cdot)$ denotes the logit of the main pathway.
Similarly, the softmax output of the hybrid pathway stemming from $B_L^m$ is given by
\begin{equation}
 q_H^m(x; \theta_m, \tau) =\text{softmax} \left (\frac{f_H^m(x ; \theta_m)}{\tau} \right), 
\end{equation}
where $f_H^m(\cdot ; \cdot)$ and $\theta_m$ denote the logit and the model parameter of the $m^\text{th}$ hybrid pathway, respectively.

\input{algorithms/proposed_method.tex}

\subsection{Knowledge distillation}
\label{sub:knowledge}

Our goal is to learn the representation of the main pathway similar to those of the hybrid pathways by using knowledge distillation.
To this end, we employ two different kinds of loss terms; one is the cross-entropy loss and the other is the knowledge distillation loss.

The cross-entropy loss of the main pathway is given by
\begin{equation}
    \mathcal{L}_L = \text{CrossEntropy}( q_L, y),
\label{eq:main_ce_loss}
\end{equation}
while the overall cross-entropy loss of the hybrid pathways is defined as
\begin{equation}
    \mathcal{L}_{H}^\text{CE} = \frac{1}{M-1} \sum_{m=1}^{M-1} \text{CrossEntropy}(q_H^m, y).
\label{eq:hybrid_ce_loss}
\end{equation}

On the other hand, we encourage the representations of individual hybrid pathways to be similar to the main branch of the local network and employ the following knowledge distillation loss additionally:
\begin{equation}
    \mathcal{L}_{H}^{\text{KL}} = \frac{1}{M-1} \sum_{m=1}^{M-1} {\text{KL}}(\tilde{q}_H^m, \tilde{q}_L),
\label{eq:kl_loss}
\end{equation}
where $\text{KL}(\cdot, \cdot)$ denotes the Kullback-Leibler (KL) divergence between two normalized vectors, and $\tilde{q}$ is the temperature-scaled softmax output using the hyperparameter $\tilde{\tau}$.

The total loss function of the proposed method is given by
\begin{equation}
\mathcal{L} = \mathcal{L}_{L} + \lambda_1 \cdot \mathcal{L}_H^{\text{CE}} + \lambda_2 \cdot \mathcal{L}_H^{\text{KL}},
\label{eq:total_loss}
\end{equation}
where $\lambda_1$ and $\lambda_2$ are hyperparameters that determine the weights of individual terms.

The local model is optimized by backpropagation based on the loss function in \eqref{eq:total_loss}.
Note that we update the model parameters in the main pathway while the blocks from the global network in the hybrid pathways remain unchanged during the local updates.

\subsection{Learning procedure}
\label{sub:learning}

Our federated learning follows the standard protocol and starts from local updates in clients.
Each client first builds a model with either pretrained or randomized parameters and updates the parameters for a small number of iterations using local data.
The updated models are sent to the server, and the server aggregates the models by a simple model averaging.
Since the communication between the server and the clients is not stable, only a small fraction of clients typically participate in each round of the training procedure.
Finally, the server broadcasts the new model given by model averaging and initiates a new round.
Note that FedMLB is a client-side optimization approach and there is no special operation in the server.
Algorithm~\ref{alg:proposed_method} describes the detailed learning procedure of FedMLB.

\input{tables/moderate_and_large_scales}

\subsection{Discussion}
\label{sub:discussion}

FedMLB maintains the knowledge in the global model while accommodating new information from the client datasets in each communication round.
This objective is similar to that of continual learning, and knowledge distillation (KD) is widely used to this end.
However, vanilla KD-based methods~\cite{lee2021preservation,yao2021local} match the representations at the logit level, which derives model updates primarily at deeper layers while the parameters in the lower layers are less affected.
This issue can be alleviated by using the layer-wise KD techniques~\cite{romero2014fitnets}, but the independent supervisions at multiple layers may lead to inconsistent and restrictive updates of model parameters.
Contrary to these two options, the proposed approach constructs separate pathways using the static network blocks of the global model and updates the representations in each block of the local model to induce the proper outputs of the network.
At the same time, FedMLB effectively distributes the workload of the network across multiple blocks for adapting to the local data, which is helpful for maintaining the representations of the global model during local iterations.

Compared to the federated learning techniques that handle client heterogeneity by employing global gradient information for the local update, the proposed algorithm has the following major advantages.
First, FedMLB does not require any additional communication overhead such as global gradient information~\cite{karimireddy2019scaffold, xu2021fedcm}. 
Note that the increase in communication cost challenges many realistic federated learning applications involving clients with limited network bandwidths.
Also, unlike~\cite{karimireddy2019scaffold, acar2020federated, li2021model}, the clients are not supposed to store their local states or historical information of the model, which is particularly desirable for the low-rate participation situations in federated learning.

Meanwhile, FedMLB incurs a moderate increase of computational cost due to the backpropagation through the additional branches.
However, it achieves impressive accuracy with fewer communication rounds compared to the baselines.
The performance of FedMLB is particularly good with a relatively large number of local iterations, which is helpful for reducing the number of communication rounds even further to achieve the target accuracy.

%% file: algorithms/proposed_method.tex
%

\begin{algorithm}[t]
   \caption{FedMLB}
   \label{alg:proposed_method}
\begin{algorithmic}
   \STATE {\bfseries Input:} {\# of clients $N$, \# of communication rounds $T$, \\ \quad\quad\quad \# of local iterations $K$, initial server model $\theta^0$}
   \FOR{$\text{each round}~t = 1, \dots ,T$}
   \STATE Sample a subset of clients $S_t \subseteq \{1, \dots, N\}$.
   \STATE Server sends $\theta^{t-1}$ to each of all clients $i \in S_t$.
   \FOR{$\text{each}~ i \in S_t,~\textbf{in parallel}$}
   \STATE $\theta_{i,0}^t \leftarrow {\theta^{t-1}}$
   \FOR{$k = {1, \dots ,K}$}
   \FOR {{\bf{each}} $(x,y)$ in a batch}
   \STATE $ q_L(x ; \tau) \leftarrow \text{softmax} \left (\frac{f_L(x; \theta_{i,k-1}^t)}{\tau} \right)$
   \STATE $q_H^m(x ; \tau) \leftarrow \text{softmax} \left (\frac{f_H^m(x ; \theta_{i,m,k-1}^t)}{\tau} \right)$, \\ \quad\quad\quad\quad\quad\ $m = 1, \dots, M-1$
   
   \ENDFOR
   \STATE $\mathcal{L}(\theta_{i,k-1}^t) \leftarrow \mathcal{L}_{L} + \lambda_1 \cdot \mathcal{L}_H^{\text{CE}} + \lambda_2 \cdot \mathcal{L}_H^{\text{KL}}$
   \STATE ${\theta_{i,k}^t \leftarrow \theta_{i,k-1}^t - \eta \nabla \mathcal{L} }$  
   
   \ENDFOR
   \STATE Client sends  $\theta_{i,K}^t$ back to the server
   \ENDFOR
   \STATE {\bfseries In server:} \\ \quad {$\theta^{t}$ = $\frac{1}{|S_t|} \sum_{i \in S_t}\theta_{i,K}^{t}$}
   \ENDFOR
\end{algorithmic}
\end{algorithm}

%% file: tables/moderate_and_large_scales.tex
%

\begin{table*}[t]
\centering
\caption{Comparisons between FedMLB and the baselines on CIFAR-100 and Tiny-ImageNet for two different federated learning settings.  For (a) moderate-scale experiments, the number of clients and the participation rate are set to 100 and 5\%, respectively, while (b) large-scale experiments have 500 clients with 2\% participation rate. The accuracy at the target round and the number of communication rounds to reach the target test accuracy are based on the exponential moving average with the momentum parameter $0.9$. The arrows indicate whether the higher ($\uparrow$) or the lower ($\downarrow$) is better. }
\label{tab:moderate_and_large_scales}
\begin{subtable}[t]{1\textwidth}
\centering
\captionof{table}{Moderate-scale with Dir(0.3): 100 clients, 5\% participation}
\scalebox{0.9}{
\begin{tabular}{lcccccccc} 
\multirow{3}{*}{Method} & \multicolumn{4}{c}{CIFAR-100} & \multicolumn{4}{c}{Tiny-ImageNet} \\ 
\cline{2-9}
 & \multicolumn{2}{c}{Accuracy (\%, $\uparrow$ )} & \multicolumn{2}{c}{Rounds (\#, $\downarrow$)} & \multicolumn{2}{c}{Accuracy (\%, $\uparrow$)} & \multicolumn{2}{c}{Rounds (\#, $\downarrow$)}  \\
& \multicolumn{1}{c}{500R} & \multicolumn{1}{c}{1000R} & 47\% & 53\% & 500R           & \multicolumn{1}{c}{1000R}    & 38\%         & 42\%                 \\ 
\cline{1-9}
FedAvg~\cite{mcmahan2017communication}  & 41.88  & 47.83  & 924 & 1000+  & 33.94 & 35.42  &  1000+& 1000+ \\
FedMLB                                              & \bf{47.39} & \bf{54.58} & \bf{488} & \bf{783} &{\bf37.20} & {\bf40.16}  & {\bf539}& {1000+} \\
\cline{1-9}
FedAvgM~\cite{hsu2019measuring}                & 46.98  & 53.24  & 515 & 936  & 36.10 & 38.36  & 794& 1000+  \\
FedAvgM + FedMLB                                             & \bf{53.02} & \bf{58.97} & \bf{349} & \bf{499} & {\bf40.93} & {\bf43.52} & {\bf380}& {\bf642}  \\
\cline{1-9}
FedADAM~\cite{reddi2021adaptive}                & 47.07  & {54.19}  & 499 & 947  & \bf{36.98} & 40.60 & 647 &  1000+ \\
FedADAM + FedMLB                                              & \bf{48.59} & \bf{58.23} & \bf{472} &\bf{645} & {35.81} & {\bf42.90}  & {\bf552}& {\bf873} \\
\cline{1-9}
FedDyn~\cite{acar2020federated}                  & 48.38  & 55.78  & 425 & 735
&37.35&41.17&573&1000+ \\
FedDyn + FedMLB                                              & \bf{57.33} & \bf{61.81} & \bf{299} & \bf{377} &{\bf43.05} & {\bf46.55}  & {\bf324}& {\bf446} \\
\cline{1-9}\end{tabular}}
\end{subtable}

\vspace{0.3cm}
\begin{subtable}[t]{1\textwidth}
\centering
\captionof{table}{Large-scale with Dir(0.3): 500 clients, 2\% participation} 
\scalebox{0.9}{
\begin{tabular}{lcccccccc} 
\multirow{3}{*}{Method} & \multicolumn{4}{c}{CIFAR-100} & \multicolumn{4}{c}{Tiny-ImageNet} \\ 
\cline{2-9}
 & \multicolumn{2}{c}{Accuracy (\%, $\uparrow$ )} & \multicolumn{2}{c}{Rounds (\#, $\downarrow$)} & \multicolumn{2}{c}{Accuracy (\%, $\uparrow$)} & \multicolumn{2}{c}{Rounds (\#, $\downarrow$)}  \\
& \multicolumn{1}{c}{500R} & \multicolumn{1}{c}{1000R} & 36\% & 40\% & 500R           & \multicolumn{1}{c}{1000R}    & 26\%         & 32\%                 \\ 
\cline{1-9}
FedAvg~\cite{mcmahan2017communication}  & 29.87& 37.48& 858& 1000+ &23.63  & 29.48  & 645 &  1000+ \\
FedMLB                           & \bf{32.03} & \bf{42.61}  & \bf{642} & \bf{800} &{\bf28.39} & {\bf33.67}  & {\bf429} & {\bf710} \\
\cline{1-9}
FedAvgM~\cite{hsu2019measuring}                & 31.80 & 40.54 & 724 & 955 & 26.75 & 33.26  & 457 & 836  \\
FedAvgM + FedMLB                           & \bf{36.21} & \bf{47.75}  & \bf{496} & \bf{636} &{\bf32.00} & {\bf37.53}  & {\bf307} & {\bf500} \\
\cline{1-9}
FedADAM~\cite{reddi2021adaptive}                & 36.07 & 47.04 & 480 & 653  & 29.65 & 35.91  & 345  & 642  \\
FedADAM + FedMLB                          & \bf{38.28} & \bf{52.68}  & \bf{442} & \bf{527} &{\bf32.14} & {\bf39.54}  & {\bf311} & {\bf524} \\
\cline{1-9}
FedDyn~\cite{acar2020federated}                   & 31.58 & 41.02 & 691 & 927  &24.35 & 29.54  & 595 & 1000+   \\
FedDyn + FedMLB                           & \bf{36.50} & \bf{49.65}  & \bf{478} & \bf{611} &{\bf30.77} & {\bf37.68}  & {\bf384} & {\bf541} \\
\cline{1-9}
\end{tabular}}
\end{subtable}
\end{table*}

%% file: sections/experiments.tex



\section{Experiments}
\label{sec:experiments}

This section demonstrates the effectiveness of FedMLB by incorporating it into various baseline algorithms for federated learning.

\subsection{Experimental setup}
\paragraph{Datasets and baselines}
We conduct a set of experiments on the CIFAR-100 and Tiny-ImageNet~\cite{le2015tiny} datasets.
To simulate non-iid data, we sample examples with heterogeneous label ratios using symmetric Dirichlet distributions parametrized by two different concentration parameters \{0.3, 0.6\}, following~\cite{hsu2019measuring}.
We maintain the training dataset sizes balanced, so each client holds the same number of examples.
For comprehensive evaluation, we employ several state-of-the-art federated learning techniques including FedAvg~\cite{mcmahan2017communication}, FedAvgM~\cite{hsu2019measuring}, FedADAM~\cite{reddi2021adaptive}, and FedDyn~\cite{acar2020federated}.
We also compare with existing regularization-based approaches such as FedProx~\cite{li2020federated}, FedLS-NTD~\cite{lee2021preservation}, and FedGKD~\cite{yao2021local}.
We choose a ResNet-18~\cite{he2016deep} as the backbone network for all benchmarks, but replace the batch normalization with group normalization as suggested in~\cite{hsieh2020non}.

\begin{figure}[t]
\centering
\includegraphics[width=\linewidth]{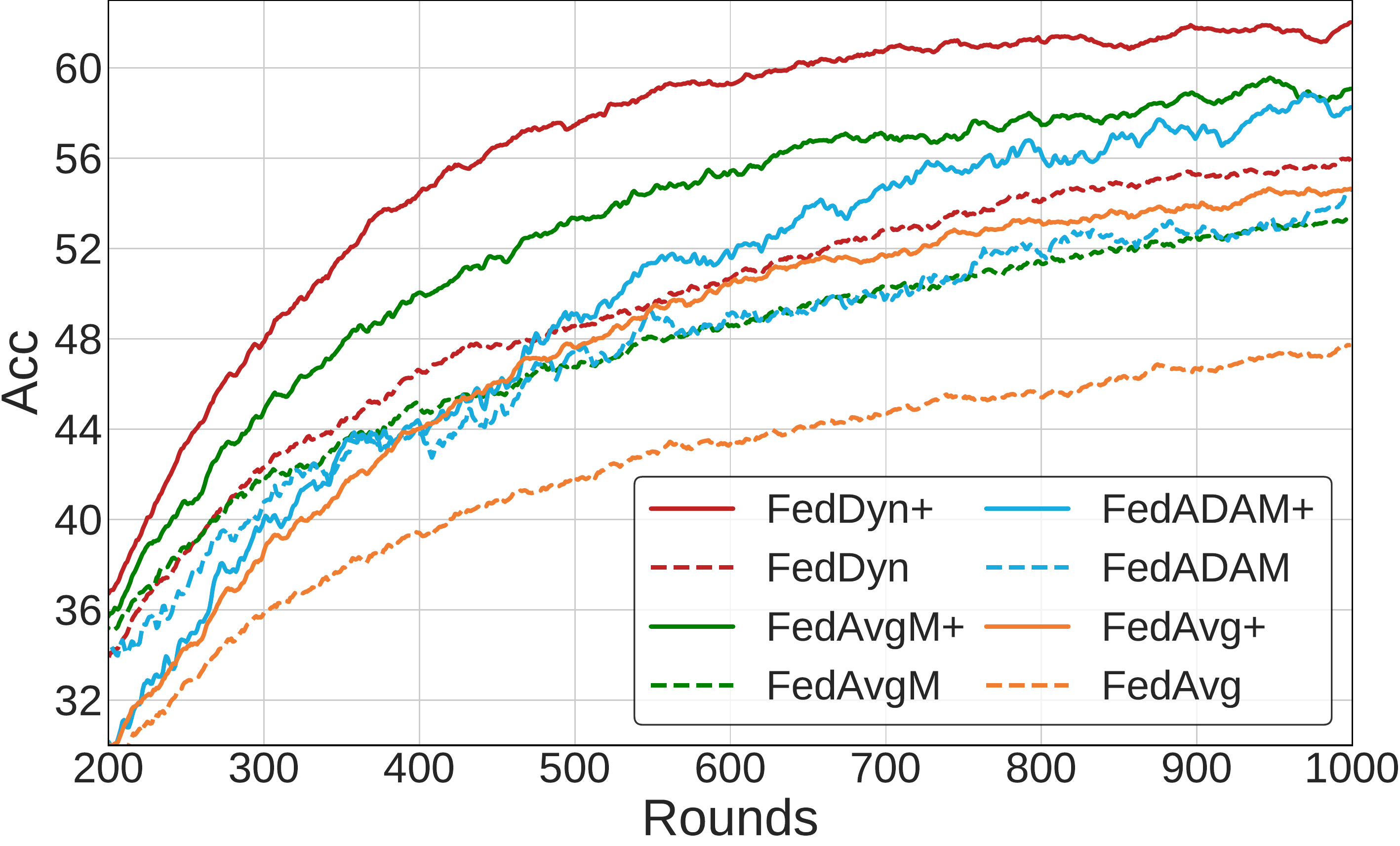}
\vspace{-0.6cm}
\caption{The convergence plots of FedMLB and other federated learning baselines on CIFAR-100.
The + symbol indicates the incorporation of FedMLB.
The number of clients, the participation rate, and the symmetric Dirichlet parameter are set to 100,  5\%, and 0.3, respectively.}
\vspace{-0.2cm}
\label{fig:main_convergence_plot_baseline}
\end{figure}

\begin{figure}[t]
\centering
\includegraphics[width=\linewidth]{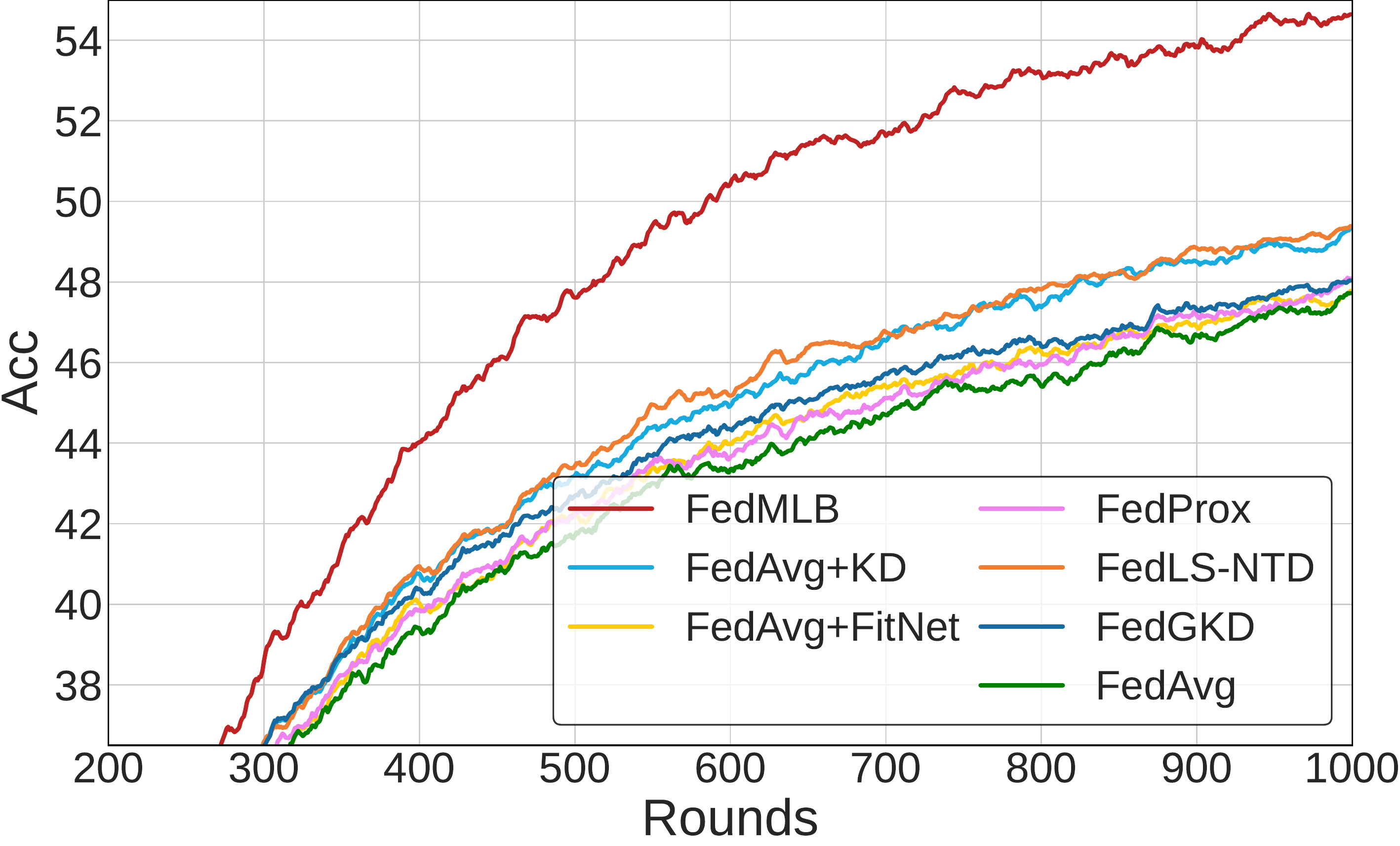}
\vspace{-0.6cm}
\caption{The convergence plots of FedMLB and other local optimization techniques on CIFAR-100.
The number of clients, the participation rate, and the symmetric Dirichlet parameter are set to 100,  5\%, and 0.3, respectively.}
\vspace{-0.2cm}
\label{fig:main_convergence_plot_distill}
\end{figure}


\input{tables/local_approaches.tex}

\input{tables/moreiteration.tex}

\input{tables/LFGN.tex}

\paragraph{Evaluation metrics}
To evaluate generalization performance of each algorithm, we use the whole test set in the CIFAR-100~\cite{krizhevsky2009learning} and Tiny-ImageNet datasets.
Since both convergence speed as well as final performance are important metrics for federated learning as discussed in~\cite{al2020federated}, we measure the performance attained at two specific rounds and the number of required rounds to achieve the desired levels of the target accuracy.
For the methods that fail to accomplish the target accuracy within the maximum communication rounds, we append a $+$ sign to the number indicating the communication round.

\paragraph{Implementation details}

We use PyTorch~\cite{paszke2019pytorch} to implement the proposed method and other baselines.
FedMLB divides its backbone network, ResNet-18, into six blocks based on the depths of its layers and feature map sizes; each of conv1, conv2\_x, conv3\_x, conv4\_x, conv5\_x, and fc layers constitutes a single block.
Following~\cite{acar2020federated, xu2021fedcm}, we adopt the SGD optimizer for local updates with the learning rate of $0.1$ for all benchmarks, except for the FedADAM whose learning rate is set to $0.01$.
We apply the exponential decay to the local learning rate with the parameter of 0.998.
There is no momentum in the local SGD but the weight decay with a factor of 0.001 is employed to prevent overfitting.
We also perform gradient clipping to stabilize the algorithms.
The number of local training epochs is set to 5, and the batch size is determined to make the total number of iterations for local updates $50$ for all experiments unless specified otherwise.
The global learning rate is $1$ for all methods except for FedADAM with $0.01$.
We list the details of the hyperparameters specific to FedMLB and the baseline algorithms in Appendix~\ref{ad:imple_details}.

\subsection{Main results}
\paragraph{FedMLB with server-side optimization techniques}
We first present the performance of the proposed approach, FedMLB, on CIFAR-100 and Tiny-ImageNet based on four federated learning baselines that perform server-side optimizations.
Our experiments have been performed on two different settings; one is with moderate-scale, which involves 100 devices with 5\% participation rate per round, and the other is with a large number of clients, 500 with 2\% participation rate.
Note that the number of clients in the large-scale setting is 5 times more than the moderate-scale experiment, which reduces the number of examples per client by 80\%.
Table~\ref{tab:moderate_and_large_scales} demonstrates that FedMLB improves accuracy and convergence speed by significant margins consistently on all the four baselines for most cases.
Figure~\ref{fig:main_convergence_plot_baseline} also illustrates the effectiveness of FedMLB when it is combined with the four baseline methods.
Note that the overall performance in the large-scale setting is lower than the case with a moderate number of clients. 
This is because, as the number of training data per client decreases, each client has even more distinct properties and is prone to drift.
Nevertheless, we observe that FedMLB outperforms the baseline methods consistently on all benchmarks.

\input{tables/hybrid_pathways}

\input{tables/avgacc_numofbranches} 
%
\paragraph{Comparisons with other local objectives}
To understand the effectiveness of FedMLB compared to other local optimization techniques, we compare our objective with the following two baselines: 1) employing the vanilla knowledge distillation for regularization (FedAvg + KD)~\cite{hinton2014distilling}), and 2) adopting knowledge distillation on block-wise features between the local and global model (FedAvg + FitNet~\cite{romero2014fitnets}).  

Table~\ref{tab:local_approaches} illustrates the outstanding performance of our multi-level branched regularization using online knowledge distillation on CIFAR-100, and Figure~\ref{fig:main_convergence_plot_distill} visualizes the convergence curves of all compared algorithms with different local objectives.
One noticeable result is that the baseline methods with knowledge distillation only achieve marginal gains or sometimes degrade accuracy.
This is partly because the predictions of the downloaded global model are not fully-trustworthy during training, especially in early communication rounds.
Therefore, merely simulating the outputs of the global model is suboptimal and hampers the learning process at the local model. 
In this respect, FedMLB is more robust to heterogeneous characteristics of clients and more flexible to learn the new knowledge in local models.
Note that, FedGKD requires 1.5 times communication costs compared to other methods since the server transmits the historical global model along with the latest model for server-to-client communication.

\subsection{Effect of more local iterations}
The increase of local iterations under a heterogeneous environment is beneficial because we can reduce the number of communication rounds between the server and clients.
Table~\ref{tab:abl_local_iterations} presents the results of FedMLB and the baselines with more local iterations, \textit{i.e.}, $K \in \{100, 200\}$, on CIFAR-100.
The results demonstrate that FedMLB outperforms the compared methods by significant margins.
An interesting observation is that the baseline methods fail to benefit from additional iterations.
This is because the increase of local iterations is prone to result in more divergence across multiple client models and eventually leads to degradation of performance.
In contrast, FedMLB consistently improves its accuracy and convergence speed substantially compared to the results with 50 iterations shown in Table~\ref{tab:local_approaches}.
Although the accuracies with 100 and 200 iterations are similar, the numbers of required iterations to achieve 40\% and 48\% are noticeably smaller with 200 iterations. 
These results imply that FedMLB handles the client drift issue effectively.

%

\input{tables/lambda_analysis.tex}
\input{tables/temp_analysis.tex}

\subsection{Analysis of auxiliary branches}

\paragraph{Effectiveness of local-to-global pathways}

Each hybrid pathway in FedMLB is composed of a set of local blocks followed by a global subnetwork.
To show the effectiveness of the current design of the hybrid pathways, we evaluate the performance of the opposite architecture design, the local model with multi-level global-to-local pathways.
As in the original version of FedMLB, the newly considered model denoted by $\text{FedMLB}_{\text{G} \rightarrow \text{L}}$ also updates the parameters in the local blocks only.
Table~\ref{tab:abl_branch_type} presents that the knowledge distillation with the hybrid pathways stemming from local blocks to global ones outperforms the opposite composition method.
The reason is that $\text{FedMLB}_{\text{G} \rightarrow \text{L}}$ constrains the output of each local blocks excessively and reduces the flexibility of the main pathway in the local network significantly.
Although the new strategy is helpful for preserving the knowledge in the global model, it interferes learning new knowledge. 

\paragraph{Effect of multi-level branches}

FedMLB employs the features from multi-level auxiliary branches to compute the cross-entropy loss $\mathcal{L}_H^{\text{CE}}$ and KL-divergence loss $\mathcal{L}_H^{\text{KL}}$.
Table~\ref{tab:path_combinations} illustrates that the pathways grafted from the lower layers are generally more helpful for accuracy gains.
Also, according to Table~\ref{tab:avgacc_numofbranches}, the accuracy of FedMLB generally improves as we increase the number of pathways.

\subsection{Effect of hyperparameters in FedMLB}

\paragraph{Ablation study for loss function}
To show the effectiveness of the cross-entropy loss of the hybrid pathways $\mathcal{L}_H^{\text{CE}}$ and the KL-divergence loss $\mathcal{L}_H^{\text{KL}}$, we conduct the comprehensive experiments by varying $\lambda_1$ and $\lambda_2$ in \eqref{eq:total_loss}, which controls the weight of each of the two loss terms.
Table~\ref{tab:lambda_sensitivity} shows that both of the loss terms contribute to performance gains while the KL-divergence loss is more critical than the cross-entropy loss in the hybrid pathways.
Note that we set $\lambda_1 = \lambda_2 = 1$ in our experiment.

\paragraph{Softmax function temperature in knowledge distillation}
The temperature parameter $\tau^{\prime}$ in \eqref{eq:kl_loss} controls the smoothness of the softmax function output for KL divergence loss, $\mathcal{L}_H^{\text{KL}}$.
Table~\ref{tab:temperature_sensitivity} presents that the performance of FedMLB is consistent with respect to the variations of $\tau^\prime$ in the two different values of the symmetric Dirichlet parameter.

%

\begin{figure}[!t]
\vspace{11pt}
  \centering
\includegraphics[width=0.95\linewidth]{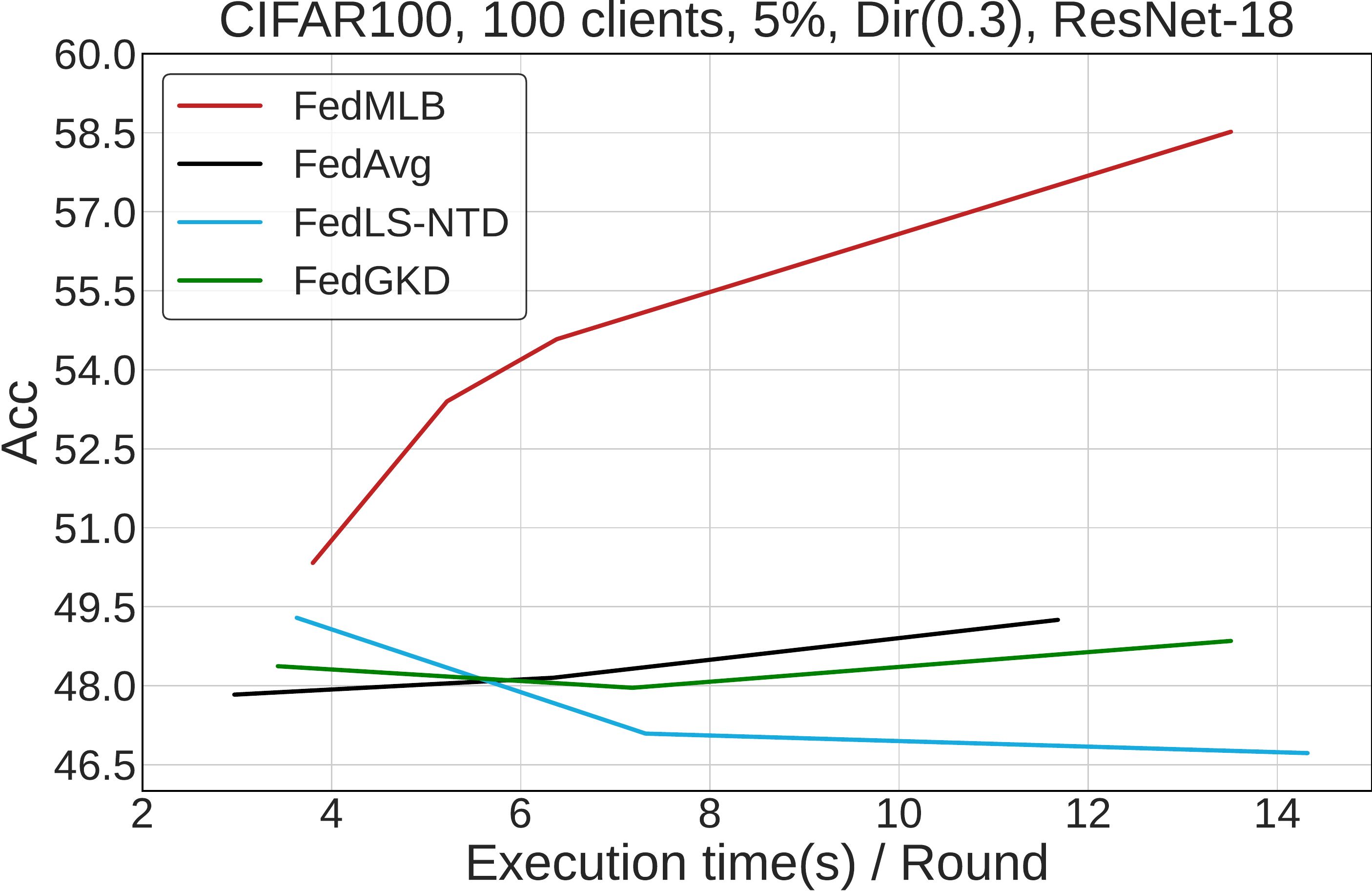}
\caption{
Performance of algorithms by varying their local computational costs controlled by the number of local iterations while maintaining the total communication costs.
Accuracies are measured after $1K$ rounds on CIFAR-100 using ResNet18.}
\label{fig:local_iterations}
\end{figure}

\subsection{Local computation overheads}
While FedMLB requires additional computation compared to the baselines under the same number of local epochs, its benefit outweighs its cost as illustrated in Figure~\ref{fig:local_iterations}.
FedMLB outperforms other methods at the same computational cost and the gap gets more significant as they have more local iterations.
An important observation from Figure~\ref{fig:local_iterations} is that FedLS-NTD and FedGKD are inconsistent with the number of local iterations compared to FedMLB and FedAVG. Also, note that the performance of FedMLB improves substantially with more iterations, which is effective for reducing the communication cost, which is critical in federated learning.
Also, Table~\ref{tab:path_combinations} presents that removing a couple of pathways from deeper layers sometimes improves accuracy, which means there still exists room for optimization in terms of computational cost.

\begin{table}[t]
\centering
\caption{Results with different CNN backbone architectures on CIFAR-100.}
\vspace{1mm}
\label{tab:diff_arch}
\scalebox{0.9}{
\begin{tabular}{lcccc}
\multirow{2}{*}{Architecture}   &  \multicolumn{4}{c}{Dir(0.3), 100 clients, 5$\%$}\\ 
&FedAvg &FedLS-NTD & FedGKD  &FedMLB\\ 
\cline{1-5}
VGG-9               &47.04&51.37&48.62&\bf{54.54}\\
MobileNet           &38.52&47.66&47.72&\bf{48.34}\\
ShuffleNet          &37.04	&39.27&38.47 &\bf{42.29}\\
SqueezeNet          &39.86	&39.56&{42.28}&\bf{42.71}\\
ResNet-18            &47.83&49.29 &47.96&\bf{54.58}\\
\cline{1-5}
\end{tabular}}
\vspace{-2mm}
\end{table}

\subsection{Experiments with other backbone models}
To confirm the generality of FedMLB with respect to backbone networks, we conduct experiments with additional architectures, which include VGG-9~\cite{simonyan2014very}, MobileNet~\cite{sandler2018mobilenetv2}, ShuffleNet~\cite{zhang2018shufflenet}, and SqueezeNet~\cite{iandola2016squeezenet}.
VGG is widely used network without skip connections while MobileNet, ShuffleNet and SqueezeNet are lightweight networks suitable for edge devices.
As for implementation, since these modern deep neural networks are typically modularized, we typically branch a pathway after a module, \textit{e.g.}, ResBlock (no branches from the layers enclosed by a skip connection).
Table~\ref{tab:diff_arch} shows that FedMLB clearly outperforms other algorithms regardless of the backbone architectures.

%% file: tables/local_approaches.tex
%
\begin{table*}[t]
\centering
\caption{Comparison between FedMLB and the baselines based on other local objectives on CIFAR-100 with two different federated learning settings.
The accuracy at the target round and the number of communication rounds to reach the target test accuracy are based on the exponential moving average with the momentum parameter $0.9$. 
}
\label{tab:local_approaches}
\vspace{0.1cm}
\scalebox{0.9}{
\begin{tabular}{lcccccccc} 
\multirow{3}{*}{Method}                                   & \multicolumn{4}{c}{Dir(0.3), 100 clients, 5$\%$}                                                                     & \multicolumn{4}{c}{Dir(0.3), 500 clients , 2$\%$}                                                                                                                                                                \\ 
\cline{2-9}
 & \multicolumn{2}{c}{Accuracy (\%, $\uparrow$ )} & \multicolumn{2}{c}{Rounds (\#, $\downarrow$)} & \multicolumn{2}{c}{Accuracy (\%, $\uparrow$)} & \multicolumn{2}{c}{Rounds (\#, $\downarrow$)}  \\
& \multicolumn{1}{c}{500R} & \multicolumn{1}{c}{1000R} & 40\% & 48\% & 500R           & \multicolumn{1}{c}{1000R}    & 30\%         & 36\%  \\ 
\cline{1-9}
FedAvg~\cite{mcmahan2017communication}              & 41.88  & 47.83  & 428 & 1000+  & 29.87& 37.48& 504& 858 \\
FedAvg + KD~\cite{hinton2014distilling}             & {42.99}  & {49.17}  & {389} & {842}   & {29.83} & {37.65}  & {505} & {859}  \\
FedAvg + FitNet~\cite{romero2014fitnets}          &  42.04 & 47.67  & 419 & 1000+   & 29.92 &37.63  &503  & 860  \\
\cdashline{1-9}
FedProx~\cite{li2020federated}                      & 42.03  & 47.93  & 419 & 1000+ & 29.28 & 36.16 & 533 & 966  \\
FedLS-NTD~\cite{lee2021preservation}                & 43.22 &  49.29 & 386 &  825 & 28.66 &35.99  & 546 & 1000+  \\
FedGKD~\cite{yao2021local}                          & 42.28  & 47.96  & 397 & 1000+  & 29.27 & 37.25 & 530&896  \\
FedMLB (ours)                                                & \bf{47.39} & \bf{54.58} & \bf{339} & \bf{523} & \bf{32.03} & \bf{42.61} & \bf{446} & \bf{642} \\
\cline{1-9}
\end{tabular}}
\vspace{-0.2cm}
\end{table*}

%% file: tables/moreiteration.tex
%
\begin{table*}[t]
\centering
\caption{Effect of more local iterations, $K = 100$ and $200$, for FedMLB and the baselines with other local objectives on CIFAR-100 with 100 clients and 5$\%$ participation rate. 
The accuracy at the target round and the number of communication rounds to reach the target test accuracy are based on the exponential moving average with the momentum parameter $0.9$. 
}
\label{tab:abl_local_iterations}
\vspace{0.1cm}
\scalebox{0.9}{
\begin{tabular}{lcccccccc} 
\multirow{3}{*}{Method}                                   & \multicolumn{4}{c}{$K = 100$ (Dir(0.3), 100 clients, 5\%)}                                                                     & \multicolumn{4}{c}{$K = 200$ (Dir(0.3), 100 clients, 5\%)}                                                                                                                                                                \\ 
\cline{2-9}
 & \multicolumn{2}{c}{Accuracy (\%, $\uparrow$ )} & \multicolumn{2}{c}{Rounds (\#, $\downarrow$)} & \multicolumn{2}{c}{Accuracy (\%, $\uparrow$)} & \multicolumn{2}{c}{Rounds (\#, $\downarrow$)}  \\
& \multicolumn{1}{c}{500R} & \multicolumn{1}{c}{1000R} & 40\% & 48\% & 500R           & \multicolumn{1}{c}{1000R}    & 40\%         & 48\%  \\ 
\cline{1-9}
FedAvg~\cite{mcmahan2017communication}              & 41.92  & 48.15  & 398 & 987  &41.45 &49.25 & 433& 897 \\
FedAvg + KD~\cite{hinton2014distilling}             & {42.58}  & {49.15}  & {385} & {905}   &42.84  & 51.48  & 381 & 751  \\
FedAvg + FitNet~\cite{romero2014fitnets}          &  39.48 & 45.80  & 531 & 1000+   & 37.17 & 43.97  & 651  & 1000+  \\
\cdashline{1-9}
FedProx~\cite{li2020federated}                      & 42.01  & 48.17  & 391 & 992 &  41.30& 48.67 & 458 &959\\
FedLS-NTD~\cite{lee2021preservation}                & 41.34 &  47.09 & 423 &  1000+ & 39.82 &46.72  & 508 & 1000+  \\
FedGKD~\cite{yao2021local}                          & 42.04  & 46.79  & 387 & 1000+  & 42.26 & 48.85 &387 & 889 \\
FedMLB (ours)                                                & \bf{52.53} & \bf{58.52} & \bf{223} & \bf{359} & \bf{53.12} & \bf{58.91} & \bf{183} & \bf{325} \\
\cline{1-9}
\end{tabular}}
\end{table*}

%% file: tables/LFGN.tex
%
\begin{table*}[t]
\centering
\caption{Ablation study results from two different compositions of the hybrid pathways on CIFAR-100 in two federated learning settings.
The accuracy at the target round and the number of communication rounds to reach the target test accuracy are based on the exponential moving average with the momentum parameter $0.9$. 
}
\label{tab:abl_branch_type}
\vspace{0.1cm}
\scalebox{0.9}{
\begin{tabular}{lcccccccc} 
\multirow{3}{*}{Method}                                   & \multicolumn{4}{c}{Dir(0.3), 100 clients, 5$\%$}                                                                     & \multicolumn{4}{c}{Dir(0.3), 500 clients, 2$\%$}                                                                                                                                                                \\ 
\cline{2-9}
 & \multicolumn{2}{c}{Accuracy (\%, $\uparrow$ )} & \multicolumn{2}{c}{Rounds (\#, $\downarrow$)} & \multicolumn{2}{c}{Accuracy (\%, $\uparrow$)} & \multicolumn{2}{c}{Rounds (\#, $\downarrow$)}  \\
& \multicolumn{1}{c}{500R} & \multicolumn{1}{c}{1000R} & 40\% & 48\% & 500R           & \multicolumn{1}{c}{1000R}    & 30\%         & 36\%  \\ 
\cline{1-9}
FedAvg~\cite{mcmahan2017communication}              & 41.88  & 47.83  & 428 & 1000+  & 29.87& 37.48& 504& 858 \\
$\text{FedMLB}_{\text{G} \rightarrow \text{L}}$     & 42.41  & 47.40  & 386 & 1000+  & 28.53 & 35.46 & 571& 1000+  \\
$\text{FedMLB}_{\text{L} \rightarrow \text{G}}$ (ours)     & \bf{47.39} & \bf{54.58} & \bf{339} & \bf{523} & \bf{32.03} & \bf{42.61} & \bf{446} & \bf{642} \\
\cline{1-9}
\end{tabular}}
\vspace{-0.3cm}
\end{table*}

%% file: tables/hybrid_pathways.tex
\begin{table}[t]
    \centering
    \captionof{table}{
    Accuracy after $1K$ rounds of FedMLB with various configurations of the hybrid pathways on CIFAR-100 in the moderate-scale setting.
    The symmetric Dirichlet parameter is set to 0.3.
    }
    \vspace{0.1cm}
    \label{tab:path_combinations}
    \scalebox{0.9}{
    \setlength{\tabcolsep}{5mm}
    \begin{tabular}{cccccc}
		\multicolumn{5}{c}{Hybrid pathway index} & \multirow{2}{*}{Acc.}  \\
                1 &2 &3 &4 &5 &  \\
                \cline{1-6}
                 \checkmark  & \checkmark  & \checkmark  & \checkmark  & \checkmark  & 54.58 \\ \cdashline{1-6}
                 \checkmark  & \checkmark  & \checkmark  & \checkmark  &             & 55.18   \\
                 \checkmark  & \checkmark  & \checkmark  &             &             & 55.03  \\
                 \checkmark  & \checkmark  &             &             &             & 54.16   \\
                 \checkmark  &             &             &             &             & 50.79  \\ \cdashline{1-6}
                             &             &             &             & \checkmark  & 48.45  \\
                             &             &             & \checkmark  & \checkmark  & 46.78  \\
                             &             & \checkmark  & \checkmark  & \checkmark  & 51.73  \\
                             & \checkmark  & \checkmark  & \checkmark  & \checkmark  & 52.65  \\ \cline{1-6}
	\end{tabular}
	}
\vspace{-0.2cm}
\end{table}

%% file: tables/avgacc_numofbranches.tex
\begin{table}[t]
    \centering
    \captionof{table}{
    Effect of the number of hybrid pathways employed for FedMLB on CIFAR-100 in the moderate-scale setting. 
The average accuracy and the standard deviation are computed over all possible combinations with the same number of hybrid pathways.
We measure the accuracy after 1$K$ rounds, where the symmetric Dirichlet parameter is set to 0.3 for  non-iid sampling.
    }
    \vspace{0.1cm}
    \label{tab:avgacc_numofbranches}
    \scalebox{0.9}{
    \setlength{\tabcolsep}{2mm}
    \hspace{-0.3cm}
    \begin{tabular}{cccccc}

                Number of pathways &1&2&3&4&5\\\cline{1-6}
                Average accuracy &47.89&51.86& 52.89& 53.69 & \bf{54.58} \\              
                Standard deviation &\ \ 3.99&\ \ 2.45& \ \ 1.78& \ \ 1.43 & - \\              

                \cline{1-6}
                             
	\end{tabular}
	}
\vspace{-0.2cm}
\end{table}

%% file: tables/lambda_analysis.tex

\begin{table}[t]
\centering
\caption{Sensitivity of FedMLB to the weights of the two regularization loss terms with respect to the accuracy after 1$K$ round on CIFAR-100 in the moderate-scale setting. The symmetric Dirichlet parameter is set to 0.3.}

\label{tab:lambda_sensitivity}
\vspace{0.1cm}
\scalebox{0.9}{

\begin{tabular}{c|cccc} 
\backslashbox{$\lambda_1$}{$\lambda_2$}          & 0 & 1& 2& 3  \\     
\cline{0-4}
0               & 47.83 &  53.27 & 54.51 & 54.21 \\
1               & 52.10 &  54.58 &55.52 & \bf{56.32}   \\     
2               & 52.37 &  53.05 & 54.28 & 54.16  \\  
3               & 49.54 & 53.00 & 54.34 & 53.57  \\  
\cline{0-4}
\end{tabular}}
\vspace{-0.2cm}
\end{table}

%% file: tables/temp_analysis.tex
%
\begin{table}[t]
\centering
\caption{Sensitivity of FedMLB to $\tau^{\prime}$ with respect to the accuracy after 1$K$ rounds on CIFAR-100 in the moderate-scale setting with two different \iffalse Dirichlet parameters.\fi values of the symmetric Dirichlet parameter.}
\label{tab:temperature_sensitivity}
\vspace{0.1cm}
\scalebox{0.9}{
\begin{tabular}{ccccccc} 
$\tau^{\prime}$                & 0.75& 1 & 1.5 & 2 & 3 & 5 \\            
\cline{0-6}
Dir(0.3)                & 53.90& 54.58  &53.87 & 53.98 & 53.60 & 53.68  \\
Dir(0.6)                & 56.10 & 56.70 & 56.72 & 56.87 & 54.98 & 54.02 \\     
\cline{0-6}
\end{tabular}}
\end{table}

%% file: sections/conclusion.tex
\section{Conclusion}
\label{sec:conclusion}

We presented a practical solution to improve the performance of federated learning, where a large number of clients with heterogeneous data distributions and limited participation rates are involved in the learning process.
To address the critical limitations, we proposed a novel regularization technique via online knowledge distillation.
Our approach employs multi-level hybrid branched networks, which reduces the drift of the representations in the local models from the feature space of the global model.
The proposed federated learning framework has the following two desirable properties; it requires no additional communication cost and spends no extra memory to store the history of local states.
We demonstrated that the proposed approach, referred to as FedMLB, achieves outstanding performance in terms of accuracy and efficiency, through a comprehensive evaluation on multiple standard benchmarks under various environments.

%% file: sections/acknowledgments.tex

\paragraph{Acknowledgments}
\label{Acknowledgments}


This work was partly supported by Samsung Electronics Co., Ltd., and by the NRF Korea grant [No. 2022R1A2C3012210, Knowledge Composition via Task-Distributed Federated Learning] and the IITP grants [2021-0-02068, Artificial Intelligence Innovation Hub; 2021-0-01343, Artificial Intelligence Graduate School Program (Seoul National University)] funded by the Korea government (MSIT).

%% file: sections/appendix.tex

\appendix

\section{Implementation Details}
\label{ad:imple_details}
For the experiments on CIFAR-100, the number of local training epochs is 5, and the local learning rate is 0.1 except for 0.01 in FedADAM.
We set the batch sizes of local updates to 50 and 10 for the experiments with 100 and 500 clients, respectively.
The parameter for learning rate decay in each algorithm is set as 0.998.
The global learning rate is 1 except for FedAdam, which adopts 0.01.
For the Tiny-ImageNet experiments, we match the total number of local iterations with other benchmarks by setting the batch sizes of local update as 100 for 100 clients and 20 for 500 clients.

There are several hyperparameters specific to each of the existing algorithms, which are typically determined to achieve the best performance by referring to the setting in the original paper.
For example, $\alpha$ in FedDyn is 0.1, $\tau$ in FedADAM is 0.001, and $\gamma$ in FedGKD is 0.2.
We select $\beta$ in FedAvgM from \{0.4, 0.6, 0.8\}, and $\beta$ in FedProx is from \{0.1, 0.01, 0.001\}.
When incorporating FedMLB into the baselines, we inherit all their hyperparameter settings.
For all experiments for FedMLB, both $\lambda_1$ and $\lambda_2$ are set to 1 while $\tau^{\prime}$ is 1.


\section{Additional Analysis}

\subsection{Auxiliary branches}

\begin{figure}[t]
\centering
\includegraphics[width=1.0\linewidth]{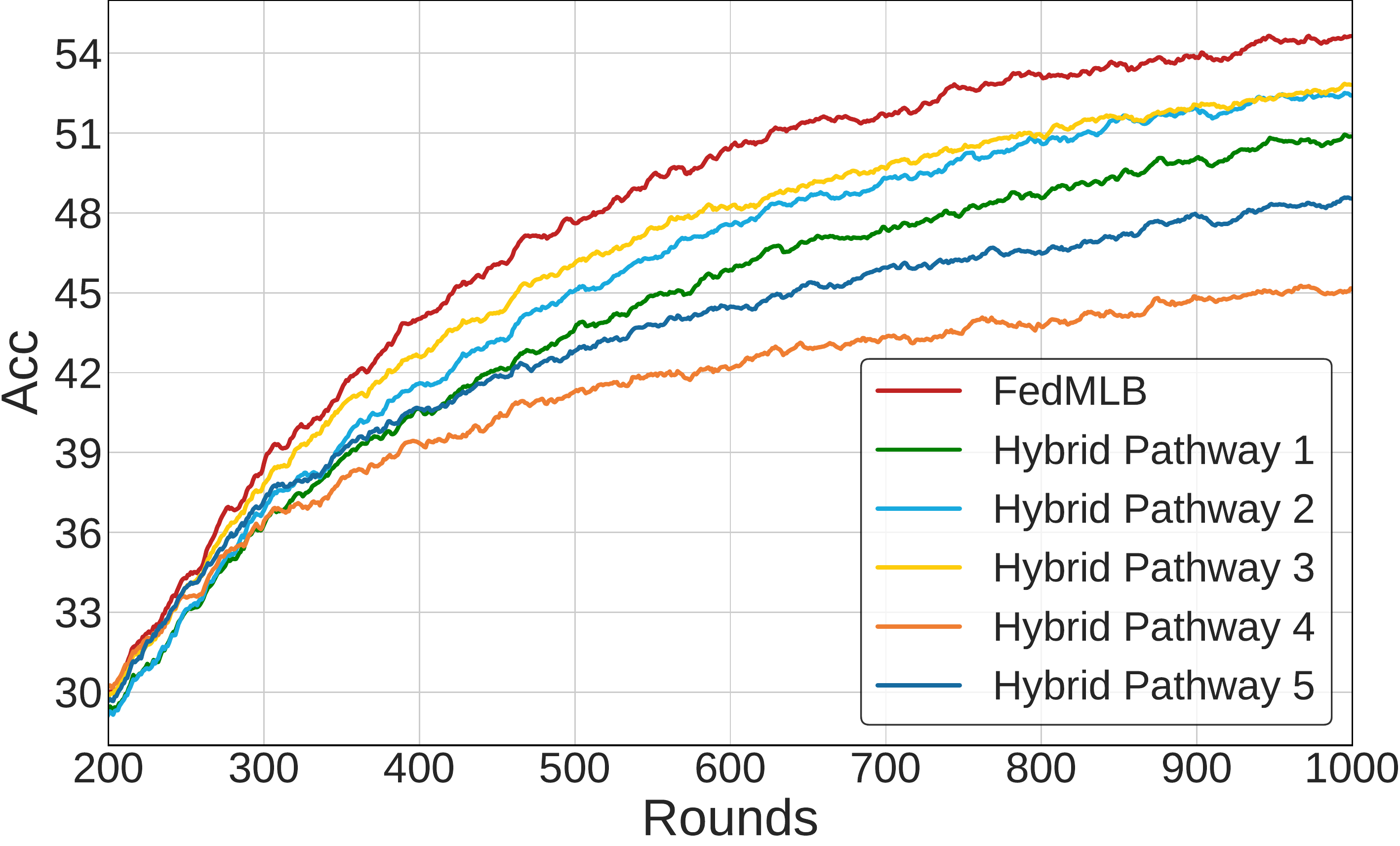}
\caption{
The benefit of each hybrid pathway to accuracy during training.
The number of clients, the participation rate, and the symmetric Dirichlet parameter are set to 100, 5$\%$ and 0.3, respectively.
The hybrid pathway index follows the notation in Figure~\ref{fig:framework}.}
\label{fig:branch_analysis}
\end{figure}

FedMLB employs the features from multi-level auxiliary branches to compute the cross-entropy loss $\mathcal{L}_H^{\text{CE}}$ and the KL-divergence loss $\mathcal{L}_H^{\text{KL}}$.
Figure~\ref{fig:branch_analysis} illustrates that the use of all auxiliary branches leads to the best performance and the proposed multi-level regularization contributes to additional performance gains.
It also implies that the branches stemming from shallower local blocks are generally more helpful, which is consistent with the results in~Table~\ref{tab:avgacc_numofbranches}.

\subsection{Level of data heterogeneity}

To demonstrate the generality of the proposed approach on the level of data heterogeneity, 
we test FedMLB with two different data partitioning strategies, which include the method based on a symmetric Dirichlet distribution with a higher concentration parameter (0.6) and the iid setting.
\cref{tab:moderate_and_large_scales_dir06,tab:moderate_and_large_scales_IID}~verify that incorporating FedMLB into four different baseline methods improves accuracy and convergence speed with large margins for most cases.

\input{tables/moderate_and_large_scales_dir06.tex}

\input{tables/moderate_and_large_scales_IID.tex}

\subsection{Convergence}

\begin{figure*}[h]
\centering
\begin{subfigure}[b]{0.48\linewidth}
\includegraphics[width=\linewidth]{figures/Addwithbaseline_5.pdf}
\caption{Dir(0.3), 100 clients, 5\% participation}
\end{subfigure}
\vspace{10pt}
\begin{subfigure}[b]{0.48\linewidth}
\includegraphics[width=\linewidth]{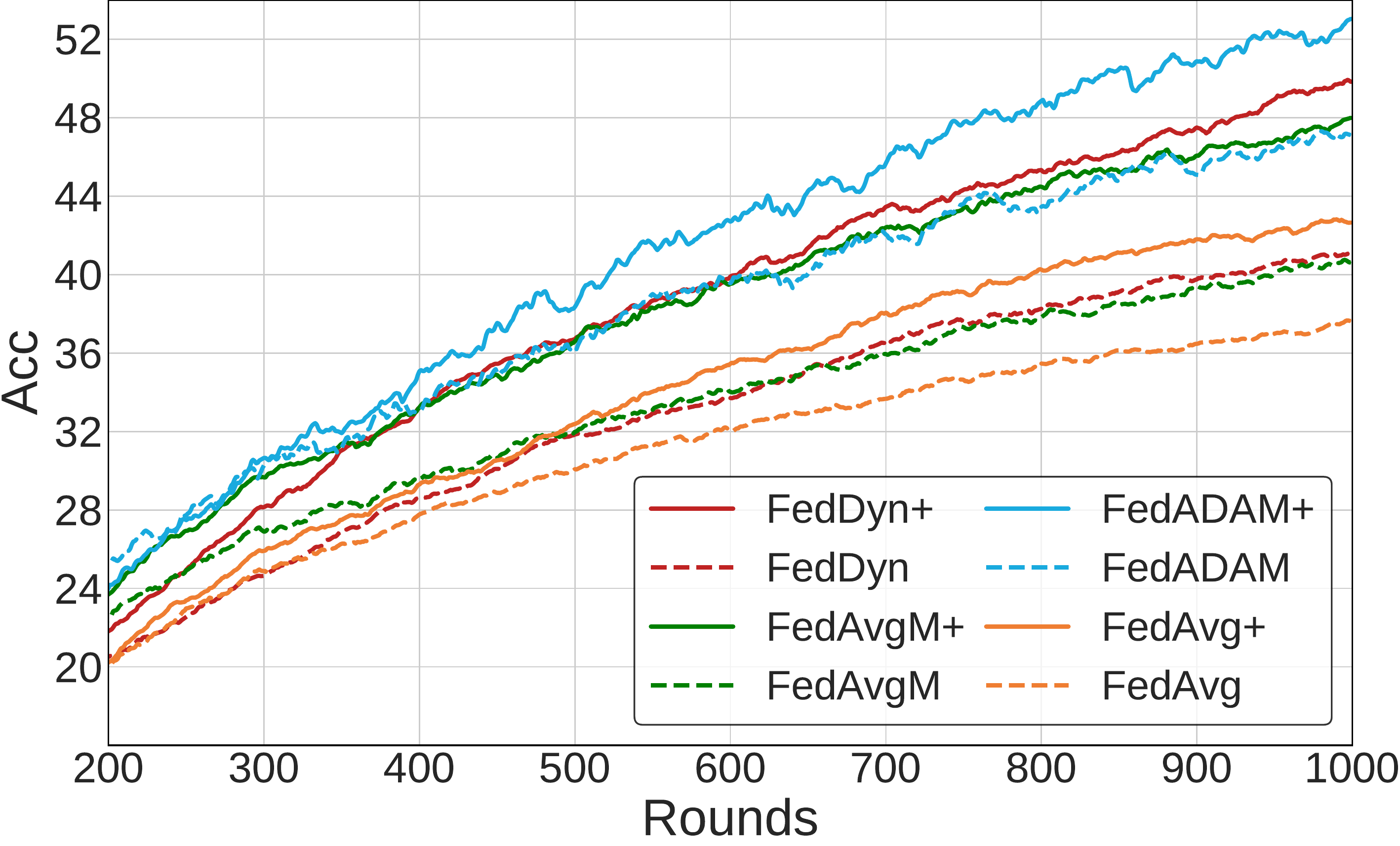}
\caption{Dir(0.3), 500 clients, 2\% participation}
\end{subfigure}
\vspace{10pt}
\begin{subfigure}[b]{0.48\linewidth}
\includegraphics[width=\linewidth]{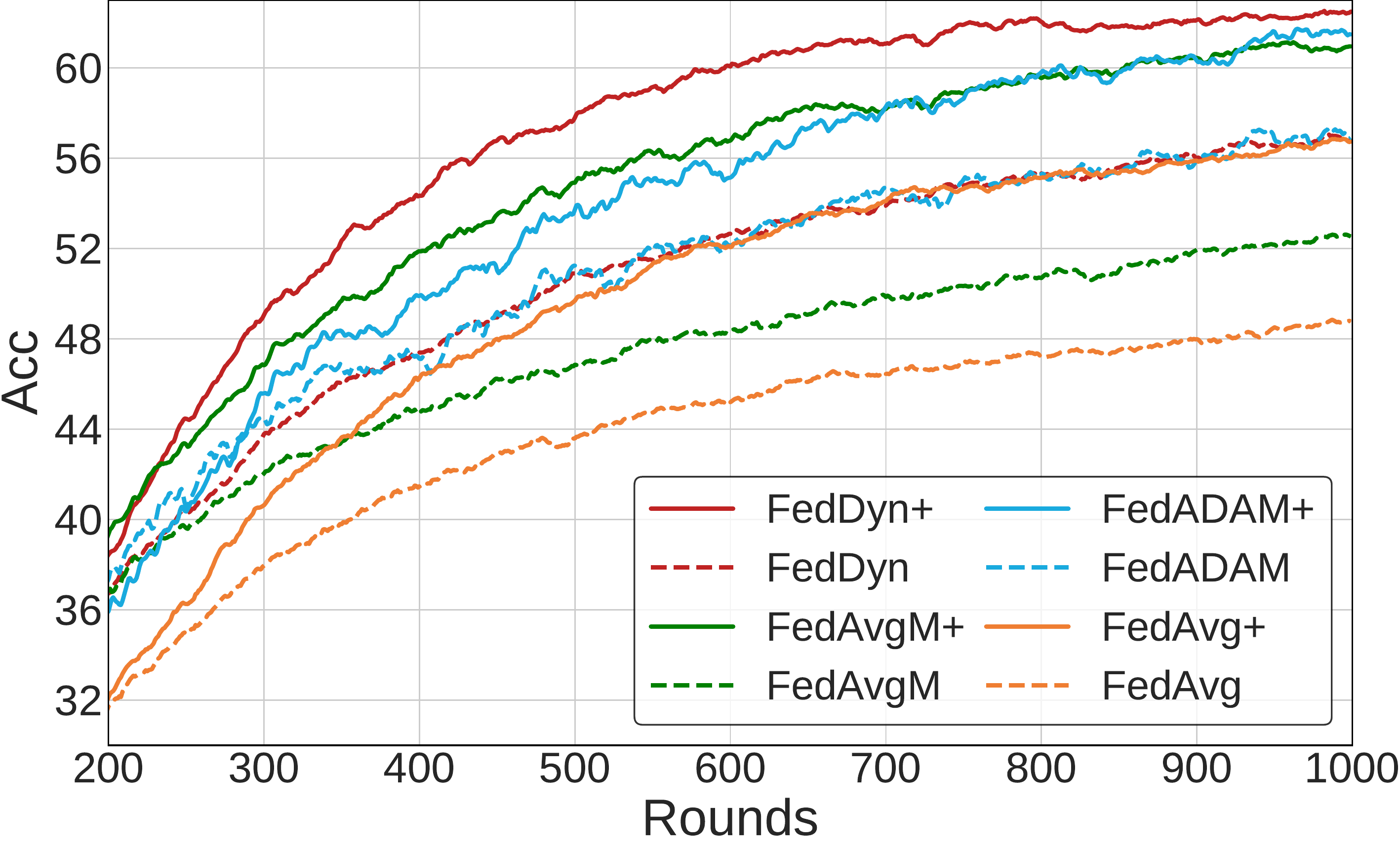}
\caption{Dir(0.6), 100 clients, 5\% participation}
\end{subfigure}
\begin{subfigure}[b]{0.48\linewidth}
\includegraphics[width=\linewidth]{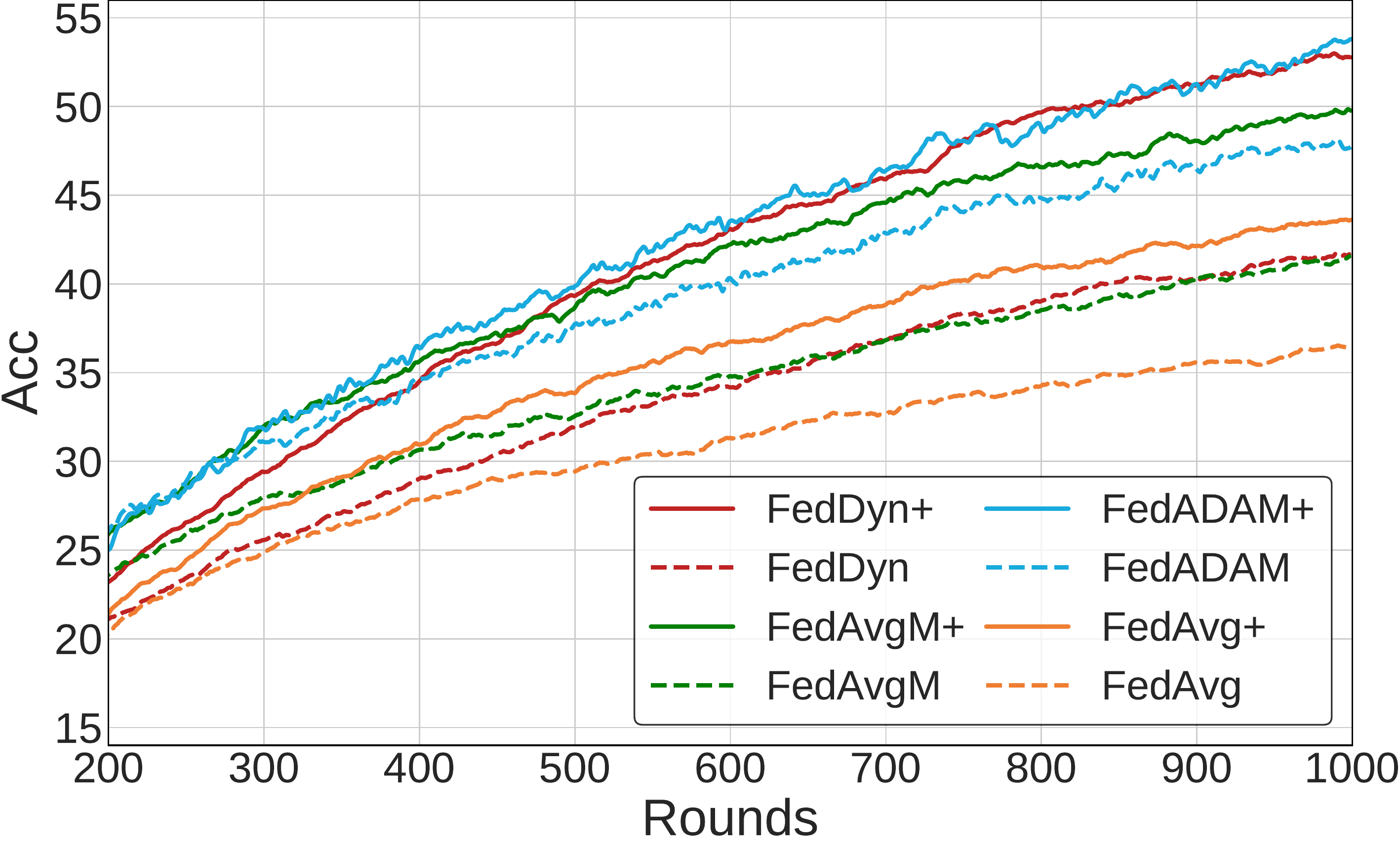}
\caption{Dir(0.6), 500 clients 2\% participation}
\end{subfigure}
\begin{subfigure}[b]{0.48\linewidth}
\includegraphics[width=\linewidth]{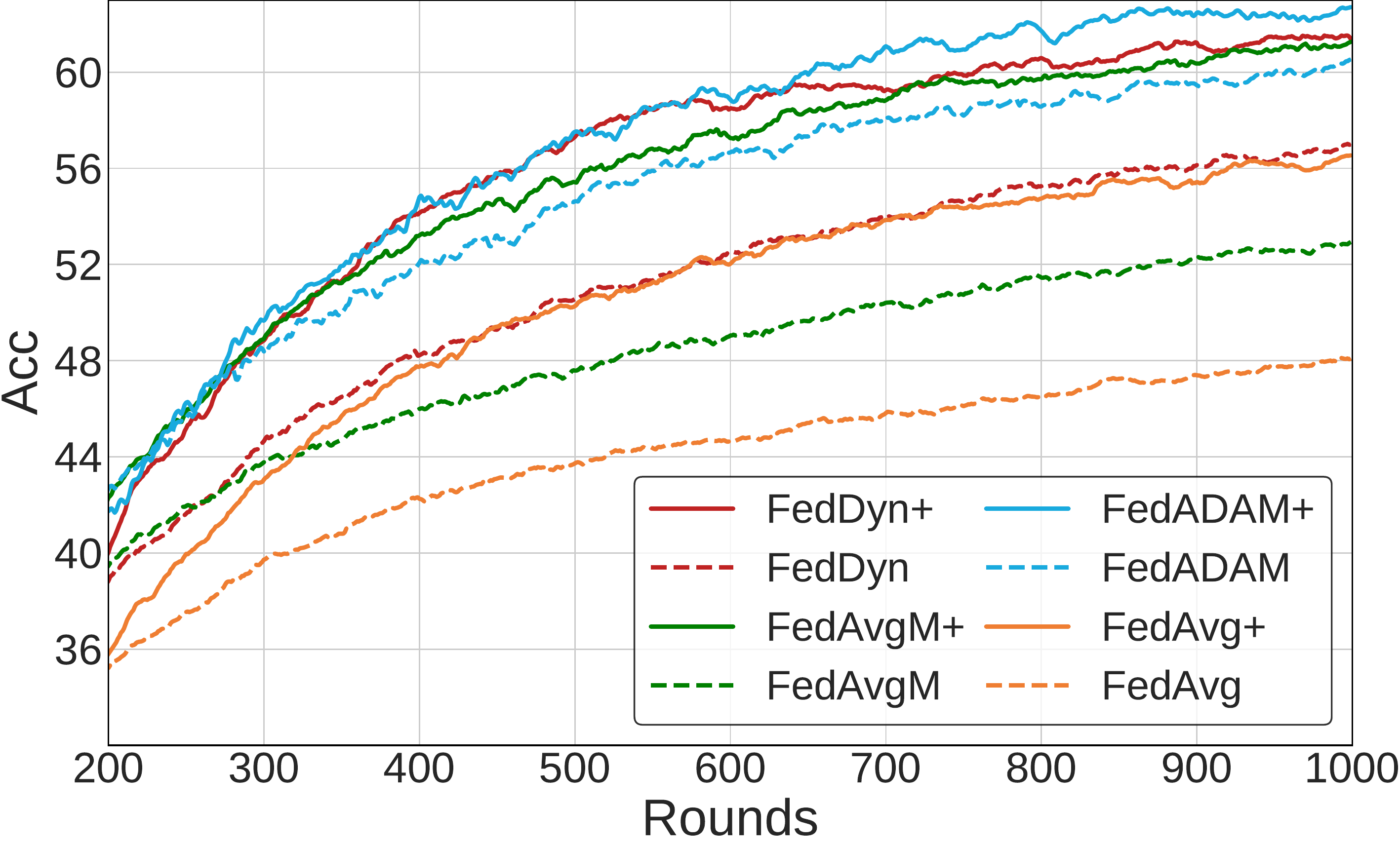}
\caption{IID, 100 clients, 5\% participation}
\end{subfigure}
\begin{subfigure}[b]{0.48\linewidth}
\includegraphics[width=\linewidth]{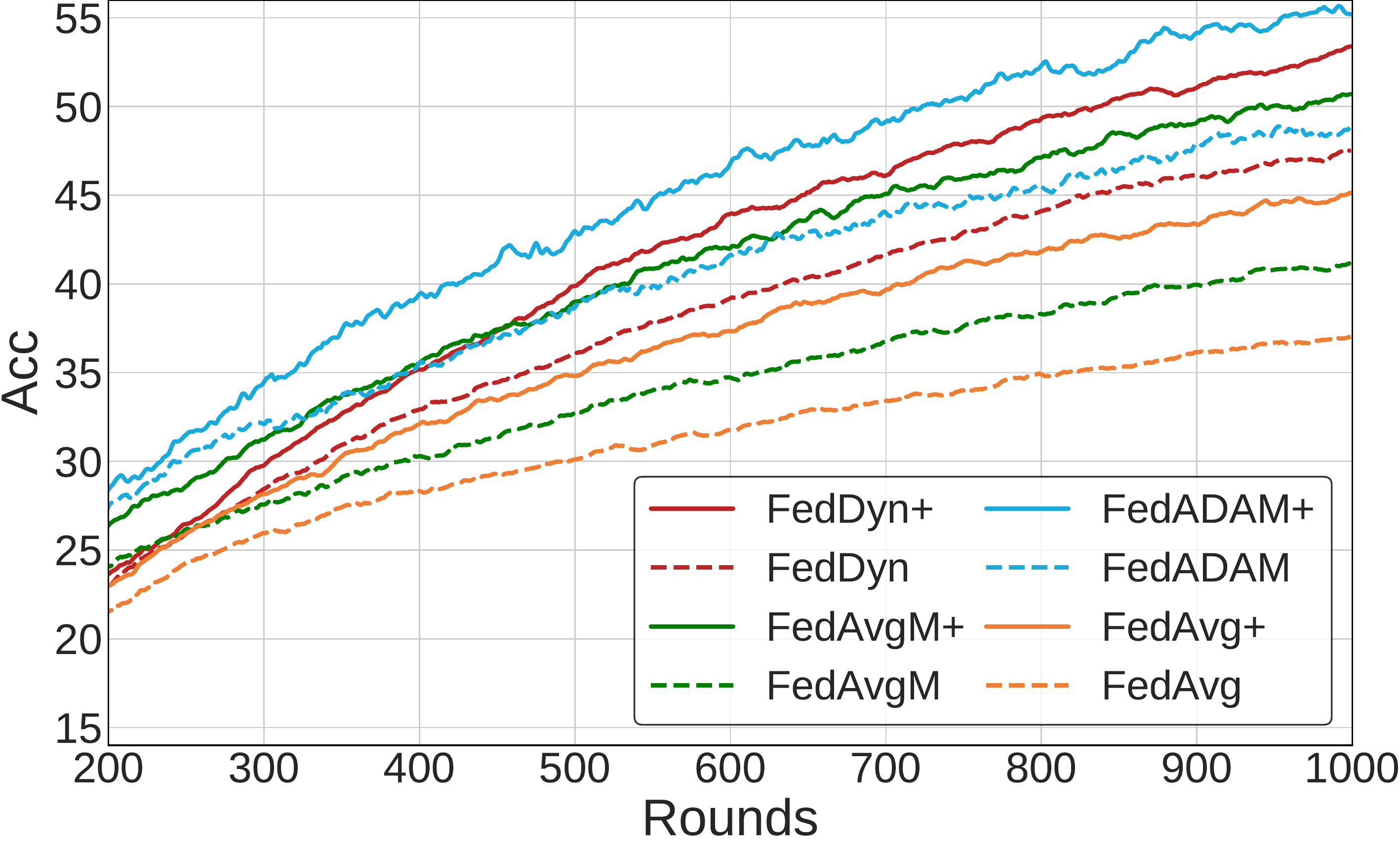}
\caption{IID, 500 clients, 2\% participation}
\end{subfigure}
\caption{
The convergence of several federated learning algorithms on CIFAR-100 in various settings. Note that + symbol indicates the incorporation of FedMLB.}
\label{fig:curve_baseline_CIFAR100}
\end{figure*}

\begin{figure*}[h]
\centering
\begin{subfigure}[b]{0.48\linewidth}
\includegraphics[width=\linewidth]{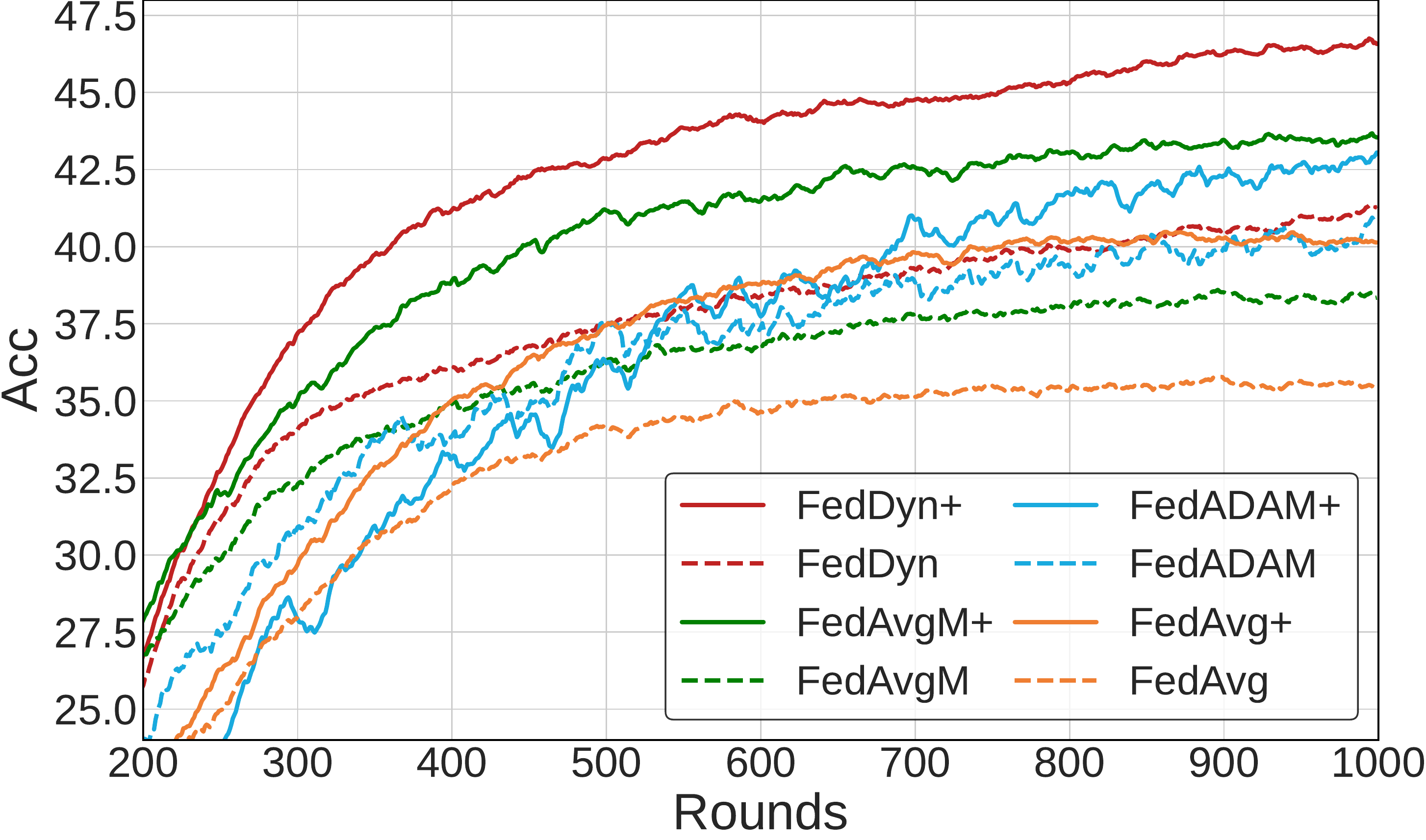}
\caption{Dir(0.3), 100 clients, 5\% participation}
\end{subfigure}
\vspace{10pt}
\begin{subfigure}[b]{0.48\linewidth}
\includegraphics[width=\linewidth]{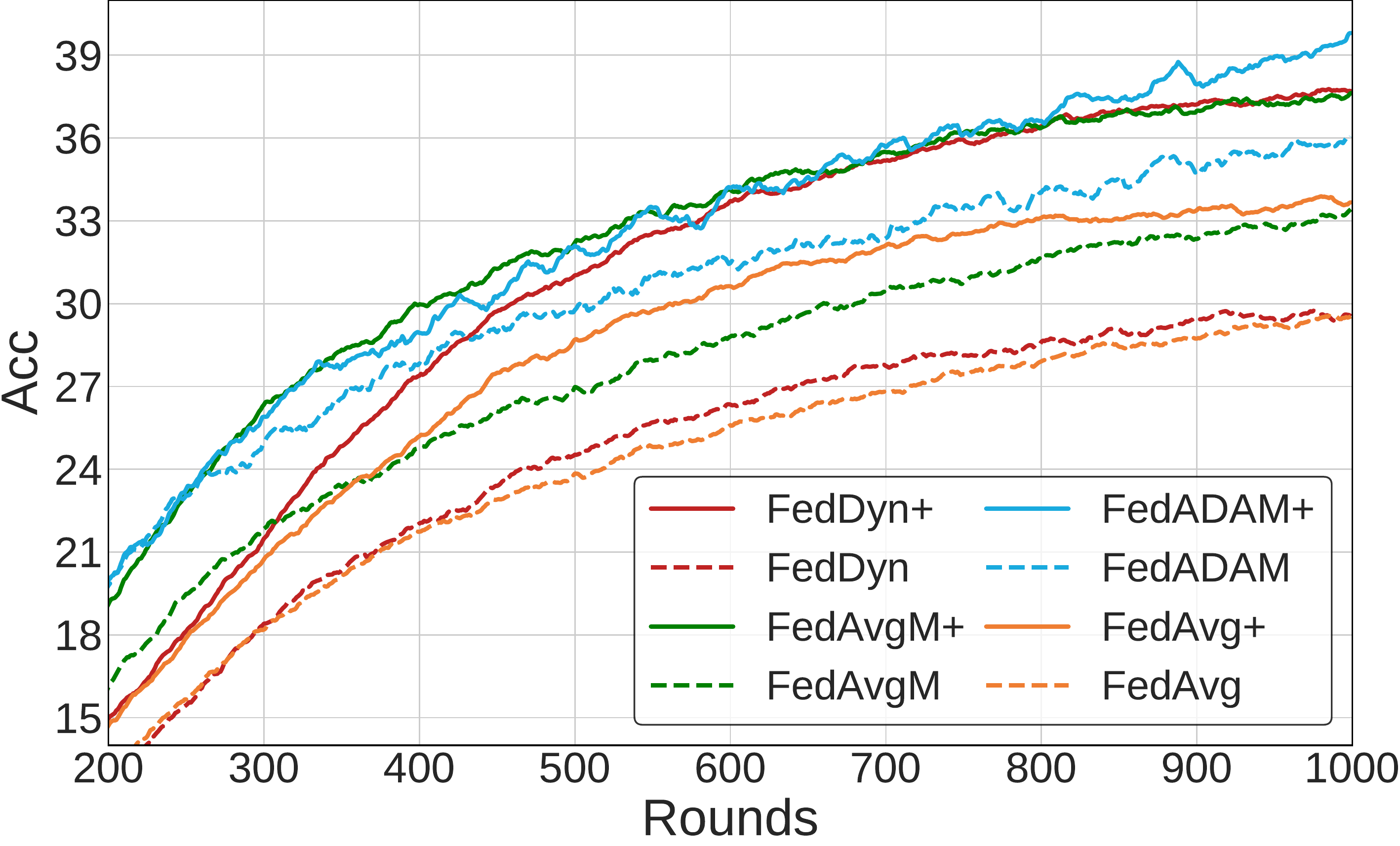}
\caption{Dir(0.3), 500 clients, 2\% participation}
\end{subfigure}
\vspace{10pt}
\begin{subfigure}[b]{0.48\linewidth}
\includegraphics[width=\linewidth]{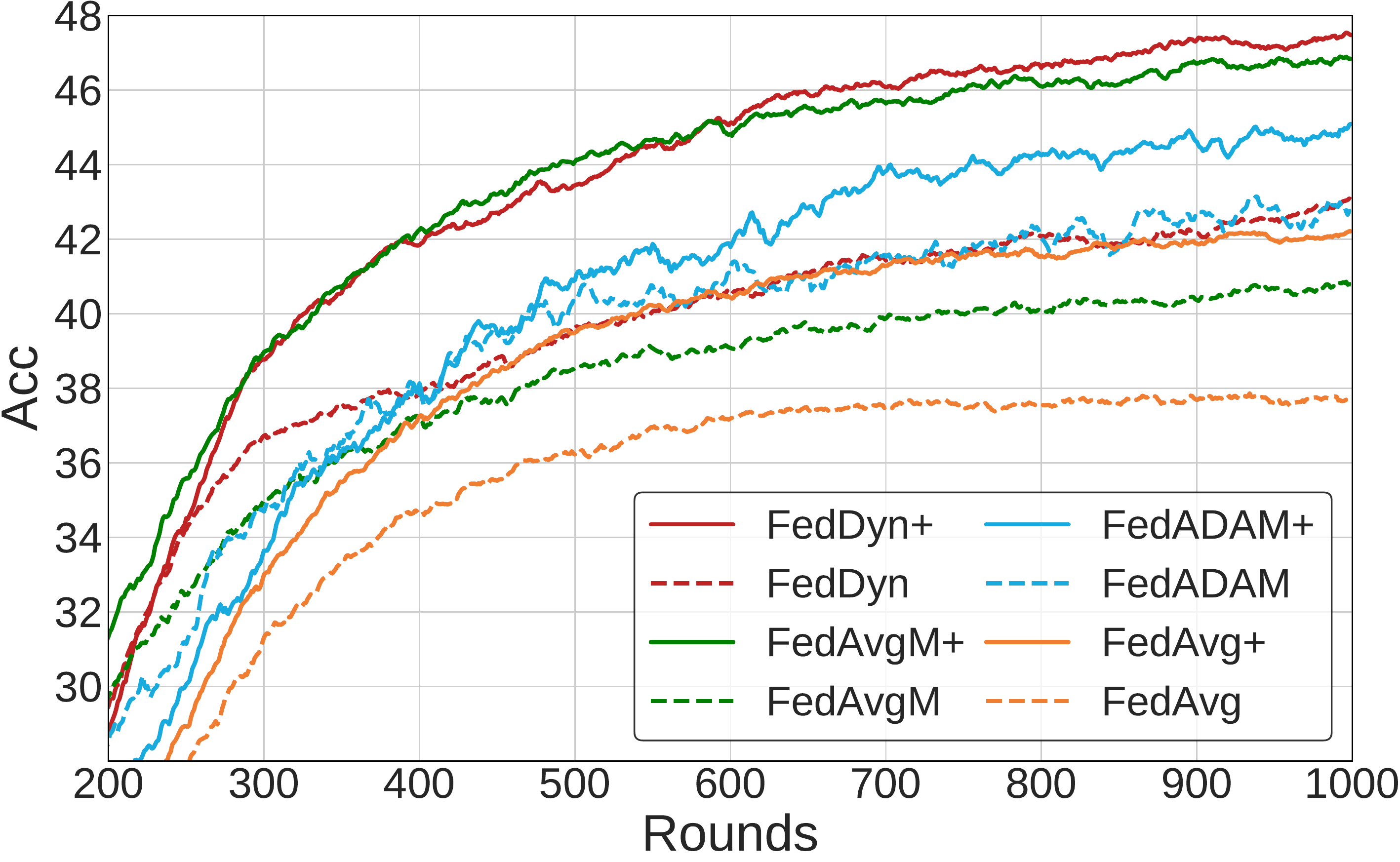}
\caption{Dir(0.6), 100 clients, 5\% participation}
\end{subfigure}
\begin{subfigure}[b]{0.48\linewidth}
\includegraphics[width=\linewidth]{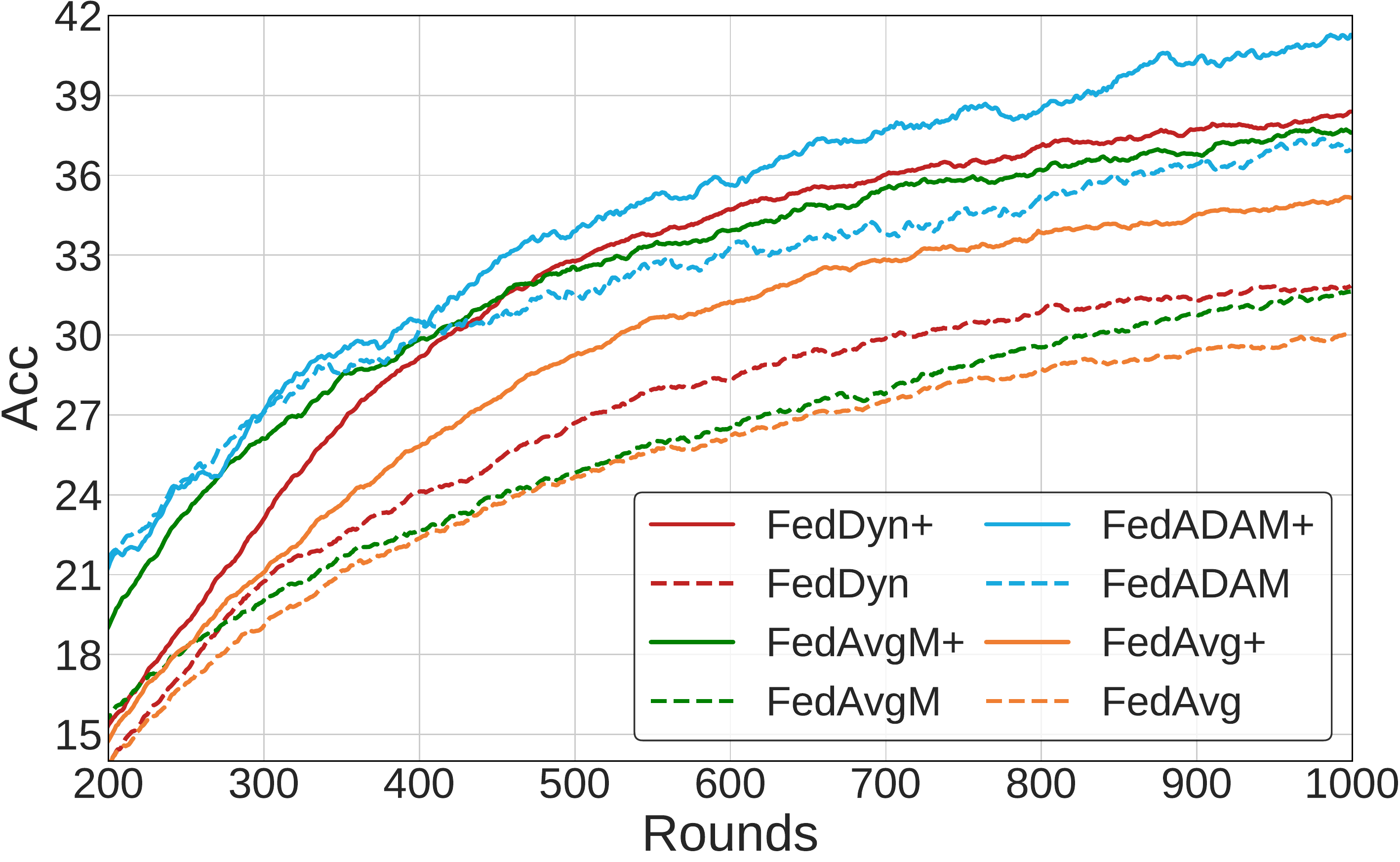}
\caption{Dir(0.6), 500 clients, 2\% participation}
\end{subfigure}
\begin{subfigure}[b]{0.48\linewidth}
\includegraphics[width=\linewidth]{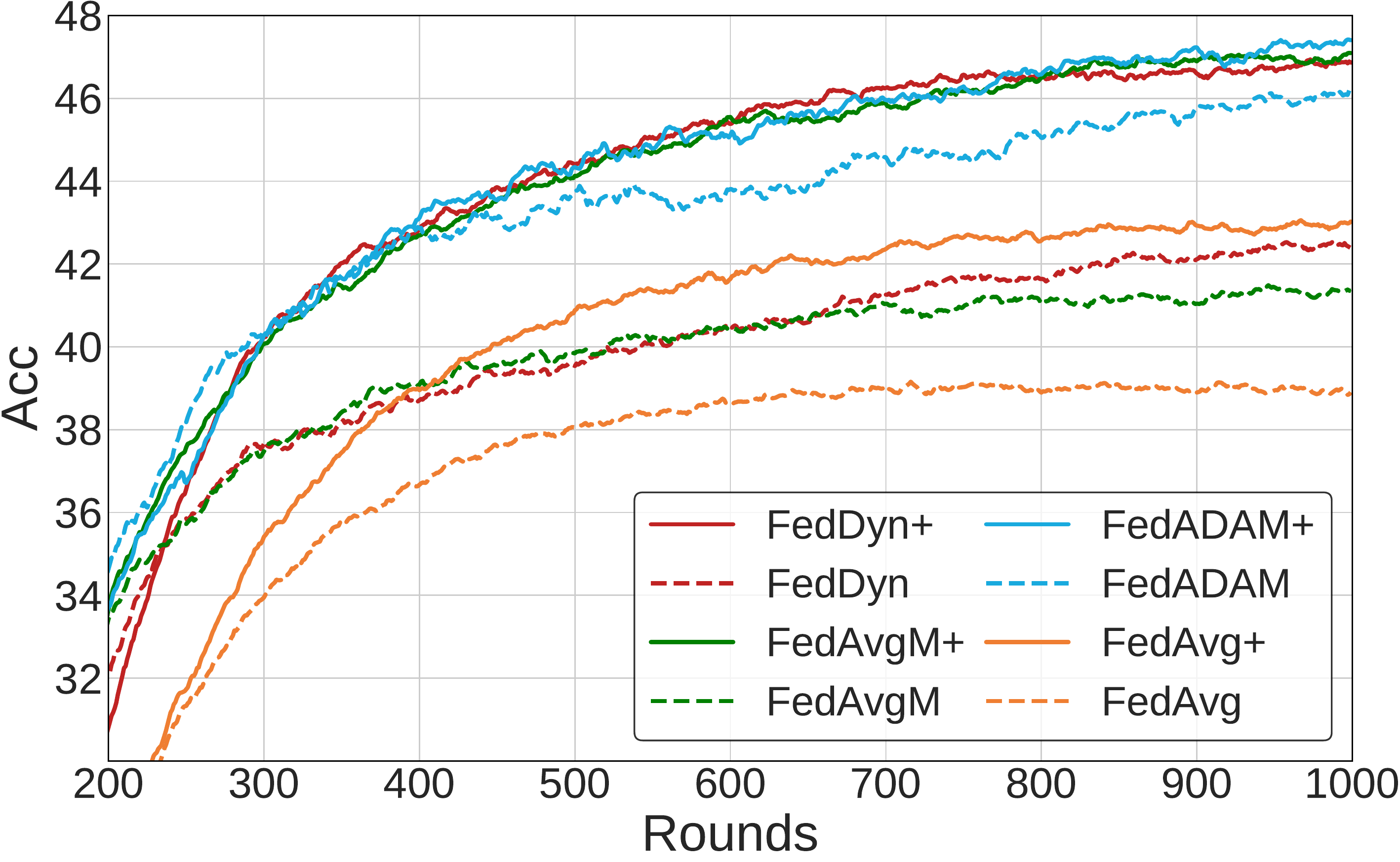}
\caption{IID, 100 clients, 5\% participation}
\end{subfigure}
\begin{subfigure}[b]{0.48\linewidth}
\includegraphics[width=\linewidth]{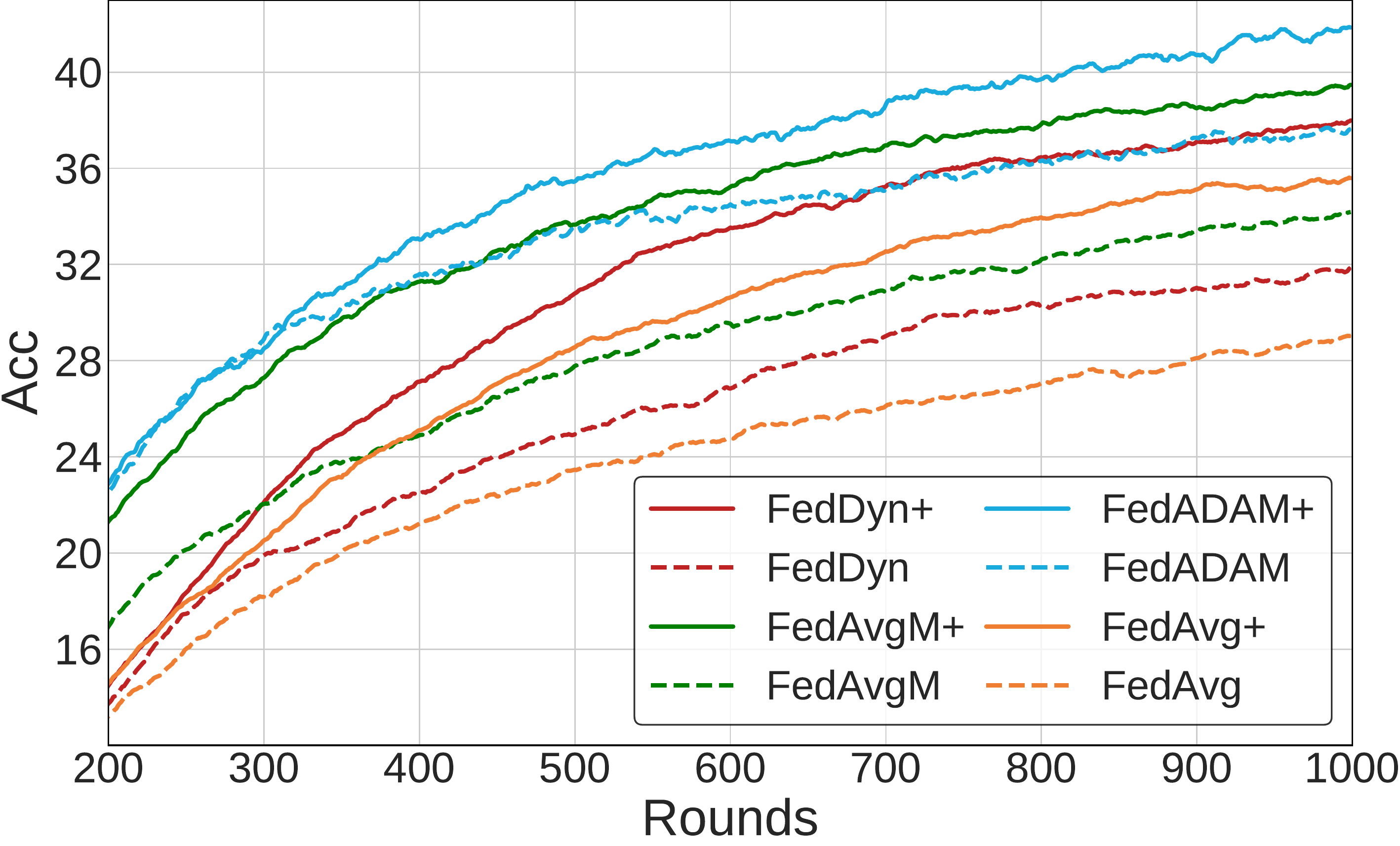}
\caption{IID, 500 clients, 2\% participation}
\end{subfigure}
\caption{The convergence of several federated learning algorithms on Tiny-ImageNet in various settings. Note that + symbol indicates the incorporation of FedMLB.}
\label{fig:curve_baseline_tiny}
\end{figure*}

To further analyze the effectiveness of the proposed method, we investigate the convergence characteristics of several algorithms including FedMLB in diverse settings, which are configured by varying the number of clients, the level of data heterogeneity, and participation rate.
The accuracies at each round on CIFAR-100 and Tiny-ImageNet are demonstrated in Figure~\ref{fig:curve_baseline_CIFAR100} and \ref{fig:curve_baseline_tiny}, respectively.
The results show that FedMLB is indeed helpful for improving accuracy throughout the training procedure, facilitating convergence.

\begin{figure*}[h]
\centering
\begin{subfigure}[b]{0.48\linewidth}
\includegraphics[width=\linewidth]{figures/main_convergence_plot_distill.pdf}
\caption{Dir(0.3), 100 clients, 5\% participation}
\end{subfigure}
\begin{subfigure}[b]{0.48\linewidth}
\includegraphics[width=\linewidth]{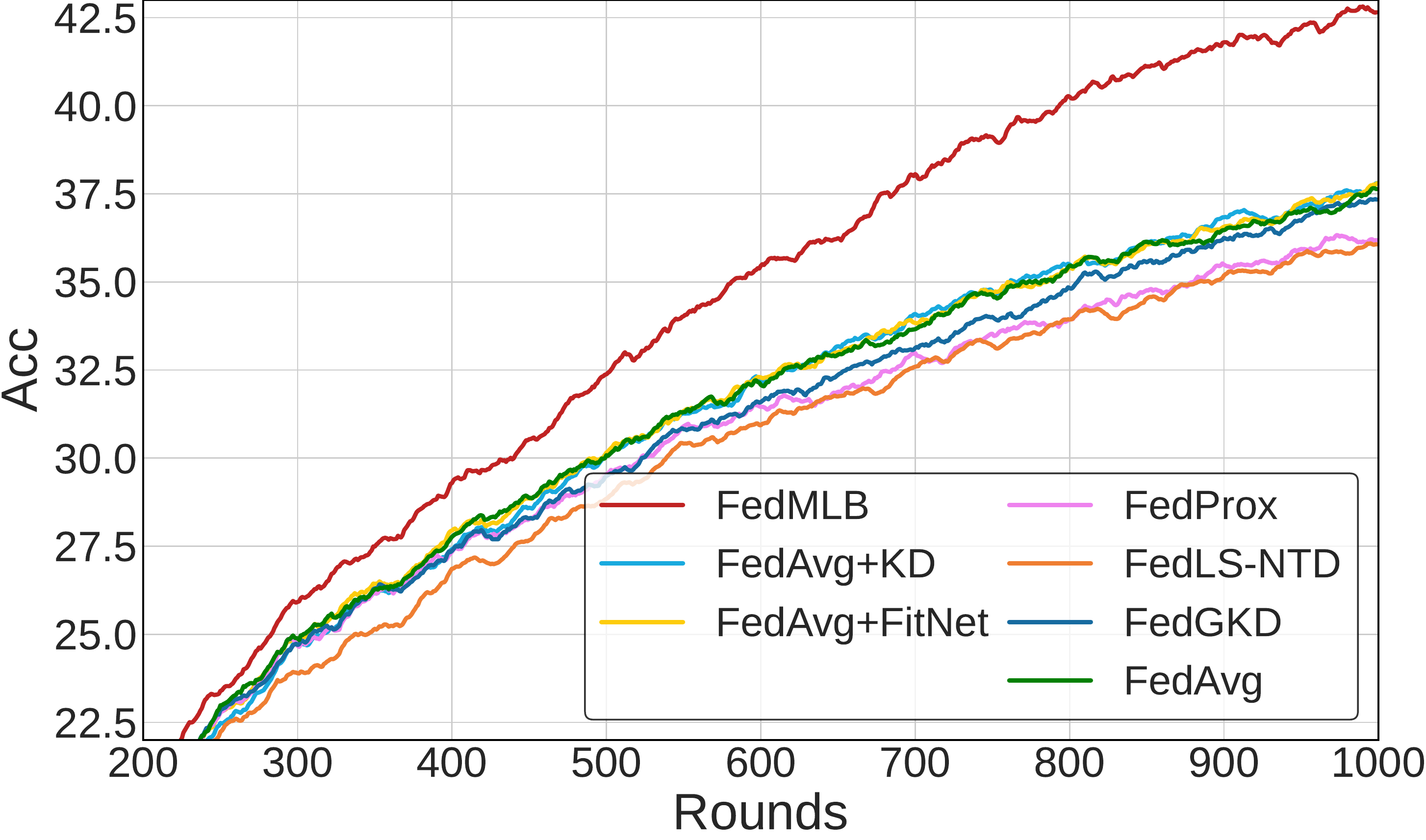}
\caption{Dir(0.3), 500 clients, 2\% participation}
\end{subfigure}
\caption{The convergence of FedMLB and other local optimization approaches on CIFAR-100.}
\label{fig:convergence_plot_distill}
\end{figure*}

\begin{figure*}[h]
\centering
\begin{subfigure}[b]{0.48\linewidth}
\includegraphics[width=\linewidth]{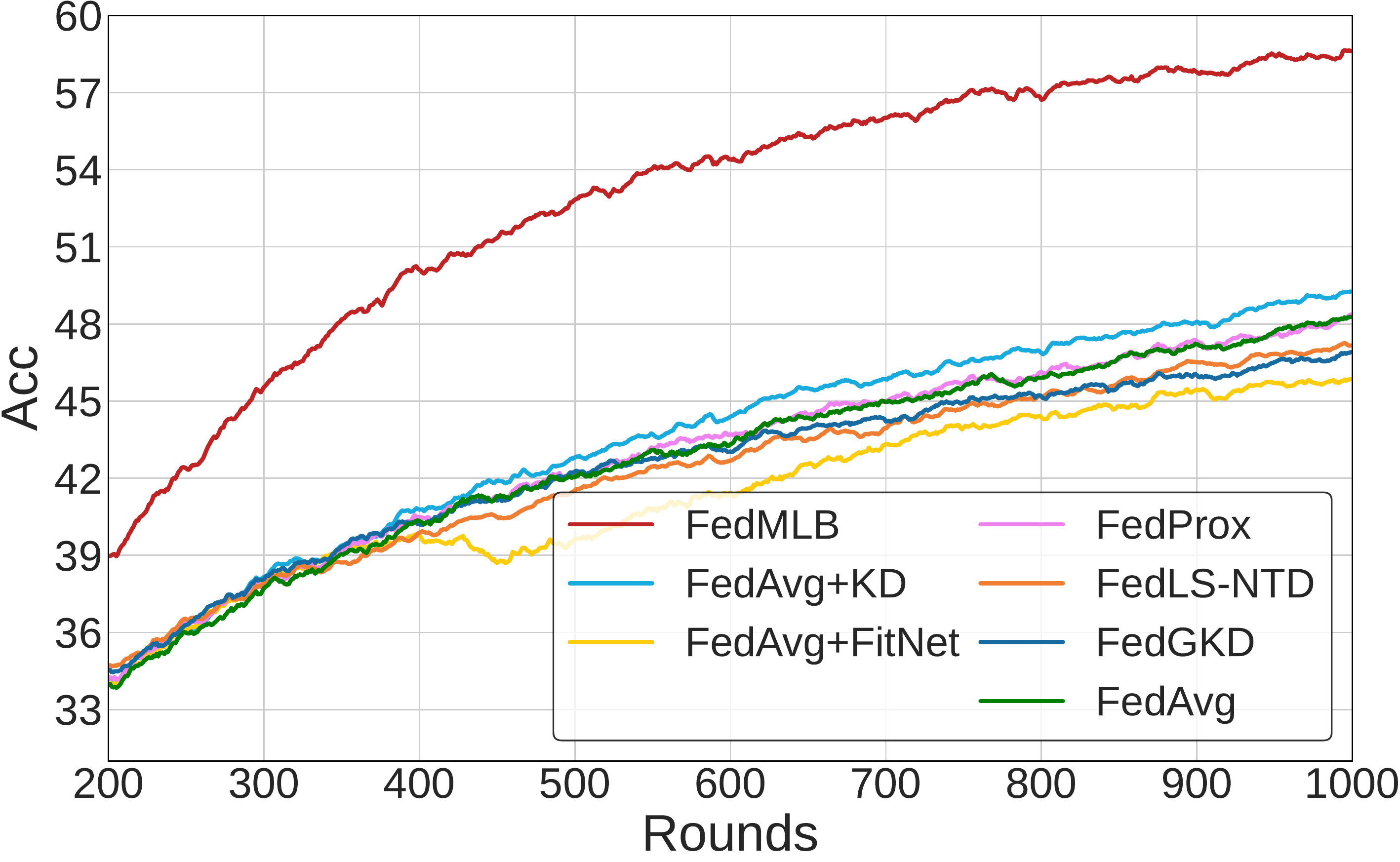}
\caption{$K = 100$ (Dir(0.3), 100 clients, 5$\%$)}
\end{subfigure}
\begin{subfigure}[b]{0.48\linewidth}
\includegraphics[width=\linewidth]{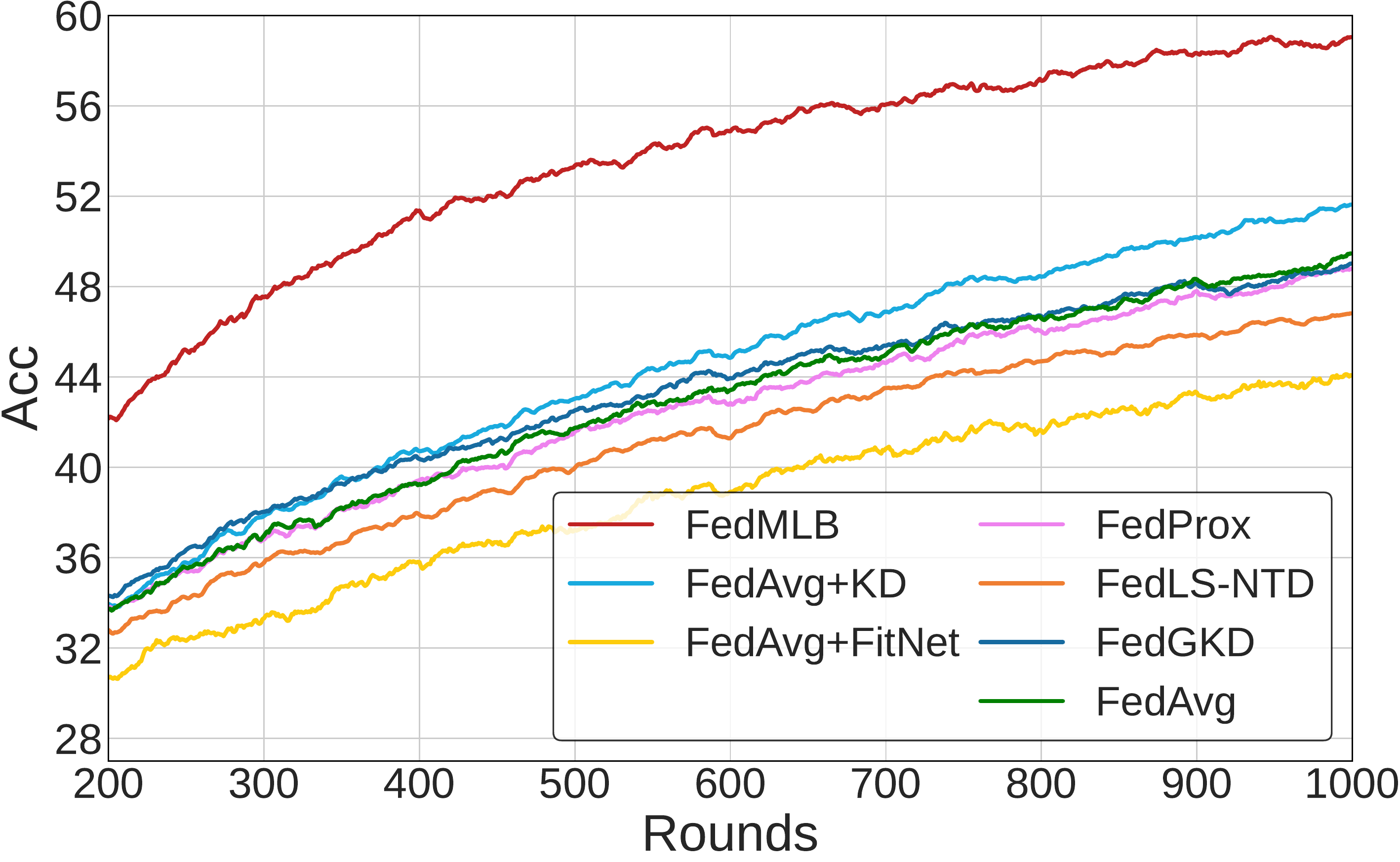}
\caption{$K = 200$ (Dir(0.3), 100 clients, 5$\%$)}
\end{subfigure}
\caption{The convergence of FedMLB and other local optimization approaches on CIFAR-100 with two different numbers of local iterations, $K$.}
\label{fig:more_local_iterations}
\end{figure*}

\Cref{fig:convergence_plot_distill} also illustrates the convergence of FedMLB in comparison to other regularization-based methods.
We observe the consistent and non-trivial improvements of FedMLB over FedAvg during training while other methods only achieve marginal gains compared to FedAvg or are even worse, especially in a more challenging condition with a less participation rate.
Furthermore, in Figure~\ref{fig:more_local_iterations} we notice that FedMLB continuously outperforms the baselines large margins even when we increase the number of local iterations, which is effective for reducing the communication cost.

%% file: tables/moderate_and_large_scales_dir06.tex
%

\begin{table*}[h!]
\centering
\caption{Comparisons of FedMLB and the baselines on CIFAR-100 and Tiny-ImageNet for two different  federated learning settings, where the symmetric Dirichlet parameter is 0.6.
The accuracy at the target round and the number of communication rounds to reach the target test accuracy are based on the exponential moving average with the momentum parameter $0.9$. The arrows indicate whether higher ($\uparrow$) or lower ($\downarrow$) is better.
}
\label{tab:moderate_and_large_scales_dir06}
\begin{subtable}[t]{1\textwidth}
\centering
\captionof{table}{Moderate-scale with Dir(0.6): 100 clients, 5\% participation}
\scalebox{0.9}{
\begin{tabular}{lcccccccc} 
\multirow{3}{*}{Method} & \multicolumn{4}{c}{CIFAR-100} & \multicolumn{4}{c}{Tiny-ImageNet} \\ 
\cline{2-9}
 & \multicolumn{2}{c}{Accuracy (\%, $\uparrow$ )} & \multicolumn{2}{c}{Rounds (\#, $\downarrow$)} & \multicolumn{2}{c}{Accuracy (\%, $\uparrow$)} & \multicolumn{2}{c}{Rounds (\#, $\downarrow$)}  \\
& \multicolumn{1}{c}{500R} & \multicolumn{1}{c}{1000R} & 48\% & 52\% & 500R           & \multicolumn{1}{c}{1000R}    & 38\%         & 42\%                 \\ 
\cline{1-9}
FedAvg~\cite{mcmahan2017communication}  & 43.43 & 48.71  & 926 & 1000+  & 36.15 & 37.72 & 1000+ & 1000+ \\
FedMLB                           & \bf{49.36} & \bf{56.70}  & \bf{465} & \bf{602}  & \bf{39.34} & \bf{42.15} & \bf{441} & \bf{917} \\
\cline{1-9}
FedAvgM~\cite{hsu2019measuring}                & 46.66 & 52.49  & 572 &  937 & 38.41 & 40.75 & 475 & 1000+ \\
FedAvgM + FedMLB                                  & \bf{54.62} & \bf{60.91} & \bf{331} & \bf{417}  & \bf{43.96} & \bf{46.82} & \bf{294} & \bf{401} \\
\cline{1-9}
FedADAM~\cite{reddi2021adaptive}                &50.81  &  56.95 & 427 & 569  & 40.13 & 42.75 & 415 & 813 \\
FedADAM + FedMLB                                  & \bf{53.34} & \bf{61.49}  & \bf{364} & \bf{467}  & \bf{40.67} & \bf{44.96} & 415 & \bf{610} \\
\cline{1-9}
FedDyn~\cite{acar2020federated}                  &50.51  &56.79   & 427 & 580  & 39.39 & 42.97 & 423 & 875 \\
FedDyn + FedMLB                                  & \bf{57.51}  & \bf{62.43}   & \bf{298}  & \bf{355}  & \bf{43.35} & \bf{47.43} & \bf{294} & \bf{412} \\
\cline{1-9}
\end{tabular}}
\end{subtable}

\vspace{0.3cm}
\begin{subtable}[t]{1\textwidth}
\centering
\captionof{table}{Large-scale with Dir(0.6): 500 clients, 2\% participation} 
\scalebox{0.9}{
\begin{tabular}{lcccccccc} 
\multirow{3}{*}{Method} & \multicolumn{4}{c}{CIFAR-100} & \multicolumn{4}{c}{Tiny-ImageNet} \\ 
\cline{2-9}
 & \multicolumn{2}{c}{Accuracy (\%, $\uparrow$ )} & \multicolumn{2}{c}{Rounds (\#, $\downarrow$)} & \multicolumn{2}{c}{Accuracy (\%, $\uparrow$)} & \multicolumn{2}{c}{Rounds (\#, $\downarrow$)}  \\
& \multicolumn{1}{c}{500R} & \multicolumn{1}{c}{1000R} & 36\% & 44\% & 500R           & \multicolumn{1}{c}{1000R}    & 26\%         & 32\%                 \\ 
\cline{1-9}
FedAvg~\cite{mcmahan2017communication}  & 29.36 &  36.36 &  966& 1000+  & 24.48 & 29.94 & 600 & 1000+ \\
FedMLB                           &\bf{33.74}  & \bf{43.53}  & \bf{571} & 1000+  & \bf{28.97} & \bf{35.08} & \bf{415} & \bf{650} \\
\cline{1-9}
FedAvgM~\cite{hsu2019measuring}         & 32.44 & 41.40  & 680 & 1000+  & 24.65 & 31.54 & 575 & 1000+ \\
FedAvgM + FedMLB                          & \bf{38.35} & \bf{49.65} & \bf{421} & \bf{690}  & \bf{32.33} & \bf{37.60} & \bf{308} & \bf{483} \\
\cline{1-9}
FedADAM~\cite{reddi2021adaptive}        & 37.33 &  47.73 & 463 &756   & 31.40 & 37.03 & \bf{286} & 533 \\
FedADAM + FedMLB                           &\bf{39.57}  & \bf{53.53}  & \bf{402} & \bf{621}  & \bf{33.71} & \bf{41.15} & 292 & \bf{444} \\
\cline{1-9}
FedDyn~\cite{acar2020federated}         & 31.63 & 41.58  & 677 & 1000+  & 26.42 & 31.80 & 485 & 1000+ \\
FedDyn + FedMLB                          & \bf{38.90} & \bf{52.72}  & \bf{440} & \bf{639}  & \bf{32.52} & \bf{38.28} & \bf{350} & \bf{712} \\\cline{1-9}
\end{tabular}}
\end{subtable}
\vspace{-0.15cm}
\end{table*}

%% file: tables/moderate_and_large_scales_IID.tex
%

\begin{table*}[h!!]
\centering
\caption{Comparisons of FedMLB and the baselines on CIFAR-100 and Tiny-ImageNet for two different iid federated learning settings. 
The accuracy at the target round and the number of communication rounds to reach the target test accuracy are based on the exponential moving average with the momentum parameter $0.9$. The arrows indicate whether higher ($\uparrow$) or lower ($\downarrow$) is better.
}
\label{tab:moderate_and_large_scales_IID}
\begin{subtable}[t]{1\textwidth}
\centering
\captionof{table}{Moderate-scale with IID: 100 clients, 5\% participation}
\scalebox{0.9}{
\begin{tabular}{lcccccccc} 
\multirow{3}{*}{Method} & \multicolumn{4}{c}{CIFAR-100} & \multicolumn{4}{c}{Tiny-ImageNet} \\ 
\cline{2-9}
 & \multicolumn{2}{c}{Accuracy (\%, $\uparrow$ )} & \multicolumn{2}{c}{Rounds (\#, $\downarrow$)} & \multicolumn{2}{c}{Accuracy (\%, $\uparrow$)} & \multicolumn{2}{c}{Rounds (\#, $\downarrow$)}  \\
& \multicolumn{1}{c}{500R} & \multicolumn{1}{c}{1000R} & 48\% & 52\% & 500R           & \multicolumn{1}{c}{1000R}    & 38\%         & 42\%                 \\ 
\cline{1-9}
FedAvg~\cite{mcmahan2017communication}  & 43.60 &  48.01 &  997& 1000+  & 37.96 & 38.92 & 504 & 1000+ \\
FedAvg + FedMLB                           &\bf{50.12}  &\bf{56.40  }&\bf{426} &\bf{584  }& \bf{40.69} & \bf{42.98} & \bf{376} & \bf{640} \\
\cline{1-9}
FedAvgM~\cite{hsu2019measuring}         & 47.43 & 52.83  & 532 & 880  & 39.79 & 41.34 & 346 & 1000+ \\
FedAvgM + FedMLB                          & \bf{55.29} & \bf{61.16}  & \bf{294} & \bf{377  }& \bf{44.02} & \bf{47.03} & \bf{271 }& \bf{382} \\
\cline{1-9}
FedADAM~\cite{reddi2021adaptive}        & 54.35 &  60.35 & 303 &416   & 43.54 &  46.12& \bf{257} & 377 \\
FedADAM + FedMLB                           &\bf{57.13}  & \bf{62.58}  & \bf{285} & \bf{363 } & \bf{44.27} &\bf{47.36} &276  &\bf{372} \\
\cline{1-9}
FedDyn~\cite{acar2020federated}         & 50.37 & 56.88  & 397 & 592  & 39.49 &  42.42& 350 & 848 \\
FedDyn + FedMLB                           & \bf{56.97 }& \bf{61.41}  & \bf{298} & \bf{366}  & \bf{44.28} & \bf{46.62} & \bf{277} & \bf{361} \\
\cline{1-9}
\end{tabular}}
\end{subtable}

\vspace{0.3cm}
\begin{subtable}[t]{1\textwidth}
\centering
\captionof{table}{Large-scale with IID: 500 clients, 2\% participation} 
\scalebox{0.9}{
\begin{tabular}{lcccccccc} 
\multirow{3}{*}{Method} & \multicolumn{4}{c}{CIFAR-100} & \multicolumn{4}{c}{Tiny-ImageNet} \\ 
\cline{2-9}
 & \multicolumn{2}{c}{Accuracy (\%, $\uparrow$ )} & \multicolumn{2}{c}{Rounds (\#, $\downarrow$)} & \multicolumn{2}{c}{Accuracy (\%, $\uparrow$)} & \multicolumn{2}{c}{Rounds (\#, $\downarrow$)}  \\
& \multicolumn{1}{c}{500R} & \multicolumn{1}{c}{1000R} & 36\% & 44\% & 500R           & \multicolumn{1}{c}{1000R}    & 26\%         & 32\%                 \\ 
\cline{1-9}
FedAvg~\cite{mcmahan2017communication}  &{29.96}& {36.93}& {903}  & {1000+} & 23.25  & 28.92 & 701 & 1000+   \\
FedAvg + FedMLB                           & \bf{34.60} &\bf{44.95}  & \bf{549} & \bf{ 935} & \bf{28.27} &\bf{35.51} &\bf{435} &\bf{ 689} \\
\cline{1-9}
FedAvgM~\cite{hsu2019measuring}         & 32.47 & 41.04  & 679 &  1000+ & 27.52 & 34.08 & 445 & 800 \\
FedAvgM + FedMLB                          &\bf{38.60} &\bf{50.52}   & \bf{418} & \bf{676}  & \bf{33.51} & \bf{39.37} & \bf{281} &\bf{444}  \\
\cline{1-9}
FedADAM~\cite{reddi2021adaptive}        &38.32  &  48.70 & 430 & 712  & 33.30 & 37.55 & 252 & 441 \\
FedADAM + FedMLB                          &\bf{42.48} &  \bf{55.24} &\bf{338}  &\bf{539}  & \bf 35.34 & \bf 41.75 & \bf{250} & \bf{384 }\\
\cline{1-9}
FedDyn~\cite{acar2020federated}         &  35.77& 47.34  & 509 &  806 & 24.79 & 31.75 &565  &1000+  \\
FedDyn + FedMLB                           & \bf{39.37} & \bf{53.11}  & \bf{429} & \bf{619}  & \bf{30.46} & \bf{37.89 }& \bf{384} & \bf{540} \\
\cline{1-9}
\end{tabular}}
\end{subtable}
\end{table*}